\documentclass{article}
\usepackage{microtype}
\usepackage{graphicx}
\usepackage{subcaption}
\usepackage{rotating}
\usepackage{multirow}
\usepackage{booktabs} 

\usepackage{hyperref}

\usepackage[preprint]{icml2026}

\usepackage{amsmath}
\usepackage{amssymb}
\usepackage{mathtools}
\usepackage{amsthm}
\usepackage{xcolor}

\usepackage[capitalize,noabbrev]{cleveref}

\theoremstyle{plain}

\theoremstyle{definition}

\theoremstyle{remark}

\usepackage[textsize=tiny]{todonotes}

\icmltitlerunning{Understanding Dynamic Compute Allocation in Recurrent Transformers}

\begin{document}

\twocolumn[
\icmltitle{Understanding Dynamic Compute Allocation in Recurrent Transformers}

\icmlsetsymbol{intern}{*}

\begin{icmlauthorlist}
\icmlauthor{Ibraheem Muhammad Moosa}{intern,yyy}
\icmlauthor{Suhas Lohit}{comp}
\icmlauthor{Ye Wang}{comp}
\icmlauthor{Moitreya Chatterjee}{comp}
\icmlauthor{Wenpeng Yin}{yyy}
\end{icmlauthorlist}

\icmlaffiliation{yyy}{Pennsylvania State University}
\icmlaffiliation{comp}{Mitsubishi Electric Research Laboratories}

\icmlcorrespondingauthor{Suhas Lohit}{slohit@merl.com}

\vskip 0.3in
]



\printAffiliationsAndNotice{* IMM performed this work as an intern at MERL. \\}  

\begin{abstract}

Token-level adaptive computation seeks to reduce inference cost by allocating more computation to harder tokens and less to easier ones. However, prior work is primarily evaluated on natural-language benchmarks using task-level metrics, where token-level difficulty is unobservable and confounded with architectural factors, making it unclear whether compute allocation truly aligns with underlying complexity. We address this gap through three contributions. First, we introduce a complexity-controlled evaluation paradigm using algorithmic and synthetic language tasks with parameterized difficulty, enabling direct testing of token-level compute allocation. Second, we propose ANIRA, a unified recurrent Transformer framework that supports per-token variable-depth computation while isolating compute allocation decisions from other model factors. Third, we use this framework to conduct a systematic analysis of token-level adaptive computation across alignment with complexity, generalization, and decision timing. Our results show that compute allocation aligned with task complexity can emerge without explicit difficulty supervision, but such alignment does not imply algorithmic generalization: models fail to extrapolate to unseen input sizes despite allocating additional computation. We further find that early compute decisions rely on static structural cues, whereas online halting more closely tracks algorithmic execution state.
\end{abstract}

\section{Introduction}

The difficulty of next-token prediction varies substantially across tokens within a sequence, depending on context, structure, and the underlying computational demands of the task. Nevertheless, most large language models (LLMs) allocate a fixed amount of computation to every token. As inference cost dominates the deployment of LLMs, this shortcoming has motivated a growing body of work on adaptive computation, including approaches that scale test-time compute by extending reasoning traces in token space~\cite{wei2022chain} or by increasing latent computation through recurrence~\cite{graves2016adaptive, dehghani2019universal, banino2021pondernet, hao2025traininglargelanguagemodels, geiping2025scaling}.

Recent advances have begun to explore \emph{token-level} adaptive computation more explicitly, allowing models to allocate different amounts of compute to different tokens~\cite{bae2025mixtureofrecursions, zhu2025scaling}. While these methods demonstrate promising efficiency--performance tradeoffs, they are typically evaluated on natural-language benchmarks using task-level metrics. In such settings, token-level difficulty is not directly observable and is entangled with architectural choices, routing mechanisms, and training heuristics. As a result, it remains unclear whether learned compute allocation policies genuinely align with underlying computational complexity, or whether they reflect incidental correlations that are difficult to interpret or validate. This limitation points to a more fundamental challenge: \emph{token-level adaptive computation makes token-level claims, yet is rarely evaluated under conditions where token-level difficulty is well-defined or testable}. Without explicit notions of token-level complexity, adaptive compute policies cannot be meaningfully interpreted, compared, or falsified. Addressing this gap requires evaluation protocols in which difficulty is controllable and observable, and experimental setups that isolate compute allocation decisions from other confounding model factors.

In this work, we study token-level adaptive computation through the lens of \emph{complexity-controlled evaluation}. We introduce a suite of algorithmic and synthetic language tasks in which difficulty can be explicitly parameterized at the token or input level, enabling direct tests of whether adaptive computation tracks true computational demands. To support controlled experimentation, we introduce \textbf{ANIRA} (Adaptive Neural Iterative Reasoning Architectures), a unified recurrent Transformer framework that enables per-token variable-depth computation while minimizing architectural confounds. ANIRA is not proposed as a performance-optimized alternative to existing models, but as a \emph{controlled experimental vehicle} for studying how and when token-level compute allocation emerges. ANIRA supports two decision mechanisms: an \emph{early-allocation} variant (ANIRA-E), which commits to a compute budget based on shallow representations, and an \emph{online-halting} variant (ANIRA-O), which makes stepwise decisions conditioned on intermediate computation. This design allows us to isolate the effect of decision timing on learned compute policies under identical training objectives and compute regularization.

Using this framework, we conduct a systematic empirical study of token-level adaptive computation across multiple dimensions, including alignment with task complexity, generalization to unseen input sizes, training dynamics, and the structure of learned allocation policies. Our findings reveal that while compute allocation aligned with complexity can emerge without explicit difficulty supervision, such alignment does not imply algorithmic generalization. Moreover, we show that early and online decision mechanisms lead to qualitatively different compute strategies, reflecting reliance on structural cues versus algorithmic execution state.

Overall, our contributions are four-fold: i) We introduce a \textbf{complexity-controlled evaluation paradigm} for token-level adaptive computation, based on algorithmic and synthetic language tasks with explicit, parameterized difficulty, enabling principled testing of whether compute allocation aligns with true computational complexity; ii) We describe \textbf{ANIRA}, a unified and controllable recurrent Transformer framework that supports per-token variable-depth computation while isolating compute allocation decisions from other architectural factors, enabling controlled comparisons between early and online decision mechanisms; iii) Using this framework, we provide a \textbf{systematic analysis} of token-level adaptive computation, showing that complexity-aligned compute allocation can emerge without explicit supervision, but does not guarantee algorithmic generalization, and that decision timing fundamentally shapes learned compute policies; and iv) We analyze the \textbf{training dynamics} of adaptive computation and identify a consistent two-phase regime—learning followed by compute reduction—providing insight into how adaptive compute policies are acquired and refined during training.

\section{Related work}

A pioneering work in adaptive computation in neural networks is ACT~\cite{graves2016adaptive} which showed that recurrent neural networks can be made adaptive to task complexity and the trained neural network can dynamically decide the number of recurrence steps at test time. The first application of such techniques to modern Transformer-based models was Universal Transformers~\cite{dehghani2019universal} which uses recurrent Transformer layers for the goal of building Universal Turing Machines with Transformer-based neural networks. PonderNet~\cite{banino2021pondernet} was proposed to overcome stability issues with ACT and was also applied to both recurrent neural networks and Transformers, and uses a halting decider at each recurrence step to decide whether to exit or to perform another recurrence step, and uses regularization to encourage early-exit. However, these models were not applied to token-level adaptivity for text generation, which is of interest to us.  

More recently recurrent/looped Transformers have been shown to be useful for various reasoning tasks. In particular, Huginn~\cite{geiping2025scaling} uses a recurrent Transformer to scale up test-time reasoning, but is not trained to perform adaptive computation. 

A parallel line of work uses adaptive computation in standard non-recurrent Transformers. This includes Depth-Adaptive Transformers~\cite{Elbayad2020Depth-Adaptive} which exits early depending at the token level, Mixture-of-Depths~\cite{raposo2024mixture}, which chooses which layers to use for computation for every token, CALM~\cite{schuster2022confident} that uses confidence-based early exiting. 

Recent work has begun to combine recurrence with token-level adaptivity for LLMs. Mixture-of-Recursions~\cite{bae2025mixtureofrecursions} explores routing strategies over recursive computation. In concurrent work, Ouro~\cite{zhu2025scaling} uses stepwise halting to scale recurrent Transformers, with evaluation primarily on natural-language benchmarks. A limitation of such evaluations is that token-level difficulty is not directly observable, making it hard to determine whether compute allocation tracks any controlled notion of complexity. In contrast, we introduce a complexity-controlled evaluation protocol in which token-level difficulty is explicitly parameterized, enabling direct tests of whether learned halting decisions align with analytically defined sources of complexity.

\section{Adaptive compute recurrent transformers}
\label{sec:method}

\subsection{ANIRA Architecture}
\label{sec:ac_architecture}

In this section, we describe the \emph{Adaptive Neural Iterative Reasoning Architectures} (ANIRA), which enables token-level adaptivity by varying the amount of computation in a recurrent Transformer core. Motivated by recurrent-depth language models~\cite{geiping2025scaling}, we treat the initial and final layers as input/output interfaces and allow variable compute only in the recurrent core, letting the model allocate more iterations to harder tokens.

ANIRA follows the Prelude--Recurrent--Coda architecture, akin to ~\citet{geiping2025scaling}. The \emph{Prelude} is a small stack of causal Transformer layers~\cite{vaswani2017attention} producing contextual token representations from the input. The \emph{Recurrent} core is a Transformer block applied for up to $D$ iterations. The \emph{Coda} is a small stack of causal Transformer layers that maps the recurrent output to next-token logits. 

Unlike~\citet{geiping2025scaling}, ANIRA is token-level adaptive. Adaptivity is learned through a separate decider module that predicts per-token exit depth, i.e., the number of recurrent iterations.\footnote{The deciders are Feedforward Networks (FFNs) matching the FFNs used elsewhere in the model.} We study two variants of this decider module, \textit{Depth Decider} that makes an early decision from the Prelude representations, and a \textit{Halting Decider} that makes online halting decisions after each iteration from the recurrent representations. 

\begin{figure}[t]
    \centering
        \centering
        \includegraphics[width=\linewidth]{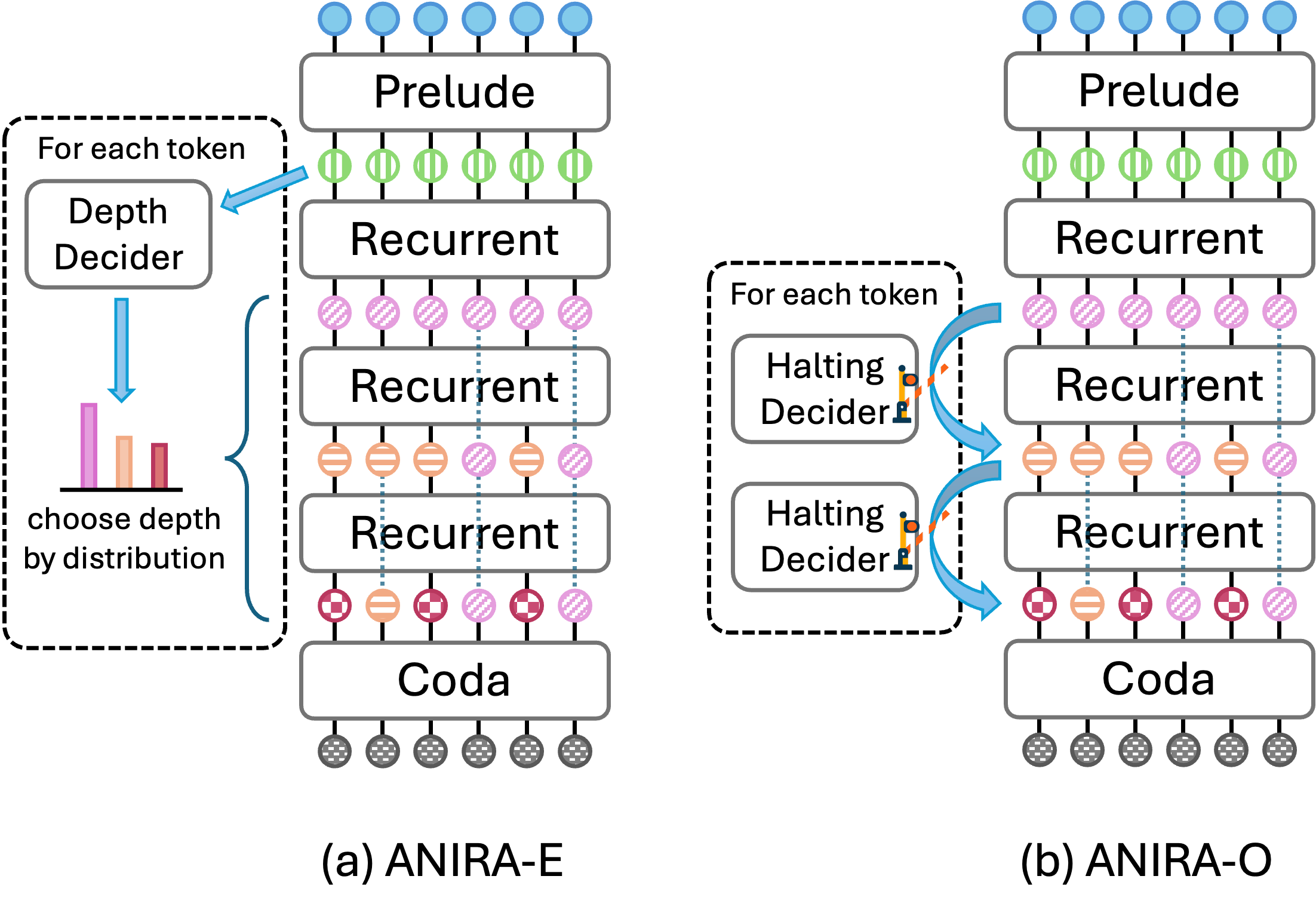}
        \caption{Two variants of ANIRA: (a) ANIRA-E: depth allocation decided from a shallow (pre-recurrence) representation. (b) ANIRA-O: online halting decisions are made for each token between each recurrent layer to decide whether to continue or halt adaptive compute. Once a token has completed its allocated number of recurrent steps, its representation is frozen and subsequent iterations act as identity mappings for that token.}
    \label{fig:anira_combined}
\end{figure}

\textbf{Architecture notation.}
Let $x_{1:T}=(x_1,\dots,x_T)$ be the sequence of input tokens. The Prelude $P(\cdot)$ is applied to the input tokens to produce the Prelude embeddings:
\begin{equation}
h^{(0)}_{1:T} \;=\; P(x_{1:T}).
\end{equation}
The recurrent block $R(\cdot)$ is applied for up to $D$ iterations,
\begin{equation}
h^{(d)}_{1:T} \;=\; R\!\left(h^{(d-1)}_{1:T}\right), \quad d=1,\dots,D.
\end{equation}
The decider allocates an exit depth $d_i^*\in\{1,\dots,D\}$ for each token
$x_i$. We define the exit representation $z_i=h^{(d_i^*)}_i, i\in\{1,\dots,T\}$ and denote
$z_{1:T}=(z_1,\dots,z_T)$ to indicate the sequence passed to the Coda.
 
Figure~\ref{fig:anira_combined} depicts the two variants that are the focus of this paper, \textit{early} depth decision (ANIRA-E) and \textit{online} halting decision (ANIRA-O).

\textbf{ANIRA-E.}
ANIRA-E makes an early depth allocation decision after the Prelude, from this shallow representation. It uses a depth decider module,   $\psi(\cdot)$, to predict a softmax distribution over the number of layers $q_i(d) = \psi(h_i^{(0)})_d, \quad d\in\{1,\dots,D\}$. 

\textbf{ANIRA-O.}
ANIRA-O makes online halting decisions, allowing the exit depth to depend on intermediate recurrent states. After computing $h^{(d)}_i$ at iteration $d$, the halting decider $\phi(\cdot)$ predicts an exit probability conditioned on the output of the $d^{th}$ iteration given by $\alpha_i^{(d)} = \phi(h_i^{(d)}) \in (0,1)$. 

These conditionals induce an exit-depth distribution via the remaining probability mass $r_i^{(d)}$, where : $r_i^{(0)} = 1$, $q_i(d) = r_i^{(d-1)}\, \alpha_i^{(d)} \text{ for } d \in \{1,\dots,D-1\}$, $r_i^{(d)} = r_i^{(d-1)}\,\bigl(1-\alpha_i^{(d)}\bigr)$, $q_i(D) = r_i^{(D-1)}$.

\textbf{Coda and prediction.}
The Coda $C(\cdot)$ takes $z_{1:T}$ as input and produces logits $o_{1:T}=C(z_{1:T})$, yielding a distribution over the predicted token, $p(x_{t+1}\mid x_{1:t})=\mathrm{softmax}(o_t)$.

\textbf{Connection to prior work.}
ANIRA-E commits to a token-specific depth decision from a shallow representation, closely related to early depth prediction in depth-adaptive Transformers \cite{Elbayad2020Depth-Adaptive} and conceptually similar to router-based recursion depth assignment \cite{bae2025mixtureofrecursions}. Unlike \citet{Elbayad2020Depth-Adaptive}, however, the decider allocates the number of iterations of a \emph{shared recurrent core}. Different from \citet{bae2025mixtureofrecursions}, it predicts an explicit per-token exit depth rather than allocating recursion via budgeted token routing/selection. ANIRA-O follows prior work on online halting-based adaptive computation \cite{graves2016adaptive,banino2021pondernet,dehghani2019universal}. However, differently, we apply online halting \emph{per token for causal language modeling} inside a shared recurrent core and enforce early-exit semantics by freezing halted token states for subsequent iterations.

\subsection{Training and Inference}
\label{sec:anira_training}
\subsubsection{Training Objective}
\label{sec:training_objective}
We train ANIRA with a cross-entropy loss $L_{\mathrm{CE}}$ and a compute regularizer $L_{\mathrm{C}}$. The overall objective is:
\begin{equation}
L = L_{\mathrm{CE}} + \gamma\, L_{\mathrm{C}}, \qquad \gamma \ge 0,
\end{equation}
where $\gamma$ controls the strength of compute regularization.

The compute regularizer encourages the model's exit-depth distribution $q_i(d)$ to match a fixed prior $p(d)$ using KL divergence: $L_{\mathrm{C}}=\frac{1}{N}\sum_{i=1}^{N}\mathrm{KL}(q_i\|p)$, where $N$ is the total number of tokens in the batch. 

We use an exponential prior over depths: $p(d) \propto b^{-d}, b\geq1$. Under this prior, $L_C$ decomposes to\footnote{When $b = 1$, the prior becomes uniform. In that case, $\mathrm{KL}(q_i\|p) = -H(q_i) + const$,
so up to an additive constant the regularizer reduces to the negative entropy term.}:
\begin{equation}
\mathrm{KL}(q_i\|p) = -H(q_i) + (\log b)\,\mathbb{E}_{q_i}[d] + \text{const}
\end{equation}
Thus, $L_{\mathrm{C}}$ penalizes expected depth guided by the 
decider distribution while discouraging degenerate depth distributions via the entropy term. $L_{\mathrm{CE}}$ is computed only over answer tokens and $L_{\mathrm{C}}$ over all tokens, since compute is incurred for both prompt and answer tokens.

\subsubsection{Passthrough Mechanism}
\label{sec:training_passthrough}
To enable efficient execution, the training computation graph must remain static. We therefore unroll the recurrent core for a fixed number of $D$ iterations, independent of per-token exit decisions. To preserve early-exit semantics, we must ensure: (i) the Coda takes as input the hidden state at each token’s selected exit depth, and (ii) the keys and values from that depth are used for subsequent attention by other tokens. We satisfy both requirements via a \emph{passthrough mechanism}.

Once an exit depth $d_i^* \in \{1,\dots,D\}$ is selected for token $x_i$ (Section~\ref{sec:training_depth_sampling}), we freeze its representation.
Let $\tilde h^{(d)}_{1:T} = R(h^{(d-1)}_{1:T})$ denote the output of the recurrent block at iteration $d$. For each position $i$ and iteration $d=2,\dots,D$, we update:
\begin{equation}
h_i^{(d)} = a_i^{(d)}\,\tilde h_i^{(d)} + \bigl(1-a_i^{(d)}\bigr)\,h_i^{(d-1)},
\qquad
a_i^{(d)}=\mathbf{1}[d \le d_i^*].
\end{equation}
It follows that $h_i^{(d)} = h_i^{(d_i^*)}$ for all $d>d_i^*$; i.e., early-exit tokens present a fixed representation to deeper recurrent iterations. With shared  parameters across iterations, this also freezes the corresponding keys and values contributed by early-exit tokens.

\subsubsection{Depth Selection at Training and Inference}
\label{sec:training_depth_sampling}
The passthrough mechanism (Section~\ref{sec:training_passthrough}) requires a discrete per-token exit depth $d_i^*\in\{1,\dots,D\}$ during training. We obtain $d_i^*$ by sampling from the exit-depth distribution $q_i(d)$. At inference time, we replace sampling with deterministic depth selection.

\textbf{ANIRA-E.}
ANIRA-E predicts a categorical distribution over depths from the Prelude representation,
$q_i(d)=\psi(h_i^{(0)})_d$, with logits $s_i\in\mathbb{R}^D$.
During training, we sample $d_i^*$ using the straight-through Gumbel--Softmax estimator~\cite{jang2017categorical}:
\begin{equation}
d_i^*=\arg\max_{d\in\{1,\dots,D\}}\frac{s_{i,d}+g_{i,d}}{\tau},
\qquad g_{i,d}\sim\mathrm{Gumbel}(0,1),
\label{eq:anirae_gumbel}
\end{equation}
and backpropagate through $\mathrm{softmax}\!\bigl((s_i+g_i)/\tau\bigr)$.
At inference, we choose the modal depth $d_i^*=\arg\max_d q_i(d)$.

\textbf{ANIRA-O.}
ANIRA-O produces conditional halting probabilities $\alpha_i^{(d)}=\phi(h_i^{(d)})$, which define an exit-depth distribution $q_i(d)$. Let $F_i(d)=\sum_{j=1}^d q_i(j)$ denote its Cumulative Distribution Function (CDF). During training, we draw $u_i\sim\mathrm{Uniform}(0,1)$ and apply inverse-CDF sampling:
\begin{equation}
d_i^*=\min_{d\in\{1,\dots,D\}: F_i(d)\ge u_i} d
\label{eq:anira_o_depth_selection}
\end{equation}
We backpropagate through the discrete choice using a straight-through estimator~\cite{bengio2013estimatingpropagatinggradientsstochastic}.
At inference, we take the median depth by setting $u_i=0.5$ in Eq.~\eqref{eq:anira_o_depth_selection} and stop once $F_i(d)\ge 0.5$ (defaulting to $d_i^*=D$ if the threshold is never crossed).

\subsubsection{Allocation-Aware KV Caching}
\label{sec:anira_inference_kv}
During autoregressive decoding, the attention computation at recurrent iteration
$d$ requires keys and values for past tokens at the same iteration. However, with a per-token early exit, a past token $i$ may have exited at iteration $d_i^*<d$, so its keys and values at iteration $d$ are not available. We therefore use an allocation-aware KV cache. For each past token position $i$, we cache KV up to its exit iteration $d_i^*$. When attention at iteration $d$ requests KV for token $i$, we retrieve the deepest cached entry, i.e., the KV at depth $\min(d,d_i^*)$.

\subsubsection{Compute and Memory Savings During Inference}
\label{sec:theory_savings}
It is easy to see that the compute and KV cache memory used by ANIRA in the recurrent block is proportional to the mean depth allocation $\bar d = \frac{1}{T}\sum_{i=1}^{T} d_i^*$ for a length-$T$ sequence. Compared to a non-adaptive model that utilizes the maximum depth $D$, ANIRA reduces compute and KV cache memory approximately according to the ratio $\bar d / D$. 

\section{Difference in compute allocation policies between ANIRA-E and ANIRA-O}
\label{sec:theoretical_analysis}

\paragraph{Activity pattern and execution cost.}
Let $a_i^{(d)}\in\{0,1\}$ indicate whether token $i$ is \emph{active} at recurrent step $d$. If $a_i^{(d)}=0$, token $i$ is frozen (no further updates), but its current representation remains visible to other tokens via attention. Let $A_d=\{i:a_i^{(d)}=1\}$. The execution-time compute cost can then be defined as $\mathrm{Cost}_{\mathrm{exec}}(\pi) \;:=\; \sum_{d=1}^D |A_d|$, where $\pi$ denotes the token-level compute-allocation policy.

\textbf{Decision cost.}
We separately account for the compute used to \emph{decide} the token-level compute-allocation policy $\pi$. Let $\mathrm{Cost}_{\mathrm{dec}}(\pi)$ denote the decision-time compute incurred by policy $\pi$.
For ANIRA-E, this is the cost of producing all per-token stopping times before running the recurrent steps using a ``small" Depth Decider head after the \textit{prelude} layer. For ANIRA-O, this includes the online per-step halting decider computation, a ``small" halting head, evaluated on active tokens. 

$\mathrm{Cost}_{\mathrm{dec}}(\pi)$ for ANIRA-E is designed to be small enough compared to $\mathrm{Cost}_{\mathrm{exec}}(\pi)$, relative to the task complexity. Without this constraint, the Depth Decider module in ANIRA-E could, in principle, reproduce the recurrent computation used by stepwise halting and thus compute the same stopping times in advance, making the comparison between ANIRA-E and ANIRA-O vacuous. This compute-limitedness constraint\footnote{We use $\mathrm{Cost}_{\mathrm{exec}}$ and $\mathrm{Cost}_{\mathrm{dec}}$ as a proxy for the expressivity of the base neural network and decider neural network, as the compute cost is directly measurable.} makes explicit that running the Depth/Halting Decider modules are intended to be \emph{much cheaper} than executing several additional recurrent steps. In our experiments, these decider heads use about $8\%$ of the total parameters.

It is easily shown that it is trivial for the halting decider in ANIRA-O to produce the same compute allocation policy as the depth decider in ANIRA-E: Fix an ANIRA-E policy $\pi^E$ with exit depth $d_i^*=\pi^E_i$ for the $i^{th}$ input token $x_i$. Making simple assumptions that the recurrent blocks in ANIRA-O can emulate identity mappings, and that the halting decider block in ANIRA-O is at least as expressive as the Depth Decider block in ANIRA-E, we can trivially define an ANIRA-O policy as $\pi^O$ by $\pi^O_i(h^{(d)}_{1:T}) = \pi^O_i(h^{(0)}_{1:T}) := \mathbf{1}\{d \ge d_i^*\}$, where the recurrent layers are simply identity mappings. Therefore, for all $d,i$:  $a_i^{(d)} = \mathbf{1}\{d \le d_i^*\}$ matches the activity pattern induced by $\pi^E$ and $\mathrm{Cost}_{\mathrm{exec}}(\pi^O)$ = $\mathrm{Cost}_{\mathrm{exec}}(\pi^E)$. 

In the reverse direction, we can similarly argue that ANIRA-E cannot emulate all the policies obtained from ANIRA-O, as the halting deciders have access to $h^{(d)}_{1:T}$ obtained from further non-trivial computation using more recurrence steps. 

Thus, we generally expect ANIRA-O to be more powerful than ANIRA-E. That is, for the same task performance, we expect ANIRA-E to make \textit{conservative} predictions of the required depth and that ANIRA-O would be able to use fewer recurrences compared to ANIRA-E. However, in practice, the training dynamics and the tasks being solved also play important roles.

\begin{figure}[t]
  \centering
  \begin{subfigure}[t]{0.49\columnwidth}
    \centering
    \includegraphics[width=\linewidth]{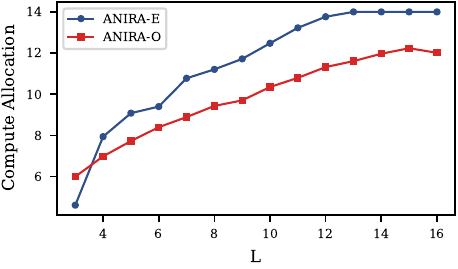}
    \caption{MANO}
    \label{fig:mano_compute_allocation_comparison}
  \end{subfigure}\hfill
  \begin{subfigure}[t]{0.49\columnwidth}
    \centering
    \includegraphics[width=\linewidth]{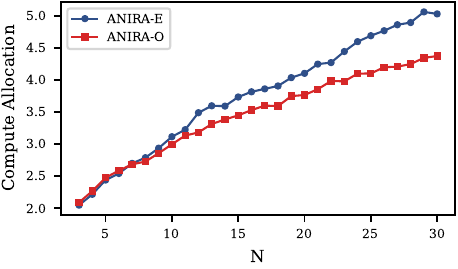}
    \caption{BREVO}
    \label{fig:brevo_compute_allocation_comparison}
  \end{subfigure}
  \caption{Task complexity vs mean depth allocation. Both ANIRA variants allocate compute consistent with task complexity. In both these cases, ANIRA-O is able to choose fewer recurrent steps for the same task performance, compared to ANIRA-E. }
  \label{fig:compute_allocation_comparison}
\end{figure}

\begin{figure*}[t]
  \centering
  \begin{subfigure}[t]{0.24\textwidth}
    \centering
    \includegraphics[width=\linewidth]{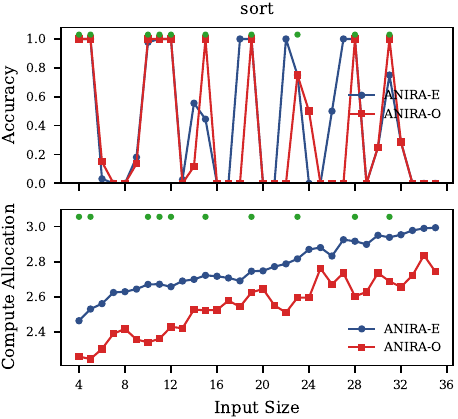}
    \caption{\textsc{sort}}
  \end{subfigure}
  \begin{subfigure}[t]{0.24\textwidth}
    \centering
    \includegraphics[width=\linewidth]{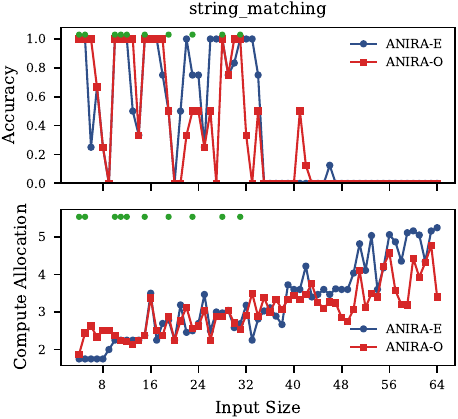}
    \caption{\textsc{string matching}}
  \end{subfigure}
  \begin{subfigure}[t]{0.24\textwidth}
    \centering
    \includegraphics[width=\linewidth]{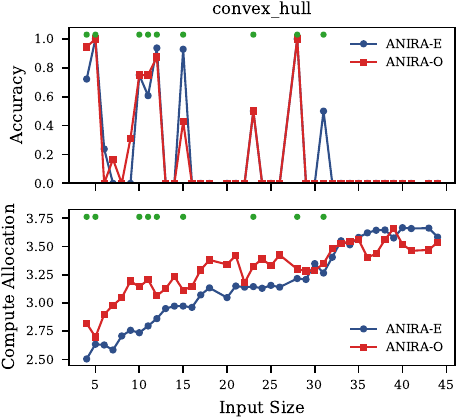}
    \caption{\textsc{convex hull}}
  \end{subfigure}
  \begin{subfigure}[t]{0.24\textwidth}
    \centering
    \includegraphics[width=\linewidth]{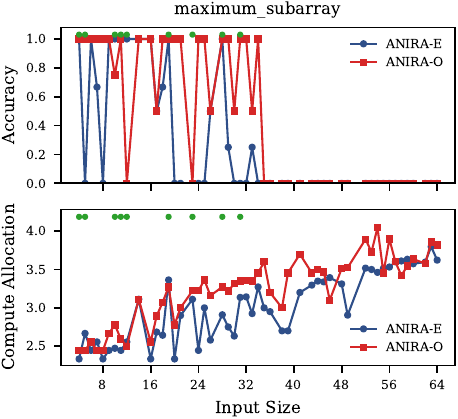}
    \caption{\textsc{maximum subarray}}
  \end{subfigure}
  \caption{CLRS task complexity vs accuracy (top) and compute allocation $\bar d$ (bottom). Green markers indicate input sizes seen during
  training. ANIRA compute allocation tracks task complexity. However, we observe that task accuracy drops sharply at input sizes not covered in training set, indicating interpolation and extrapolation failure.
  }
  \label{fig:clrs_selected}
\end{figure*}

\begin{table}[t]
\centering
\caption{Spearman correlation between per-token expected depth and complexity proxies on LANO generations.}
\small
\setlength{\tabcolsep}{6pt}
\begin{tabular}{lcc}
\toprule
Complexity Proxy & ANIRA-E & ANIRA-O \\
\midrule
Parse-space expansion        & +0.666 & +0.456 \\
Parse convergence            & -0.264 & -0.111 \\
Addition operations          & +0.618 & +0.463 \\
Multiplication operations    & +0.645 & +0.490 \\
\bottomrule
\end{tabular}
\label{tab:lano_spearman_expected_correlations}
\end{table}
\section{Complexity Controlled Evaluation Protocol}
\label{sec:evalutaion_protocol}
A central question for token-level adaptive models is whether \textit{the decider learns to allocate compute consistent with task complexity}. We design complexity-controlled evaluation protocol with algorithmic tasks where the task complexity can be rigorously defined and synthetic language where suitable token-level prediction complexity proxies can be designed. This controlled setting lets us test alignment between compute allocation and a known notion of difficulty, which is otherwise unobservable for natural language tokens.
 
\subsection{Algorithmic Tasks}
\label{sec:algorithmic_tasks}
We use the tasks \textsc{MANO} and \textsc{BREVO}  from
the Physics of Language Models benchmark suite~\cite{allen-zhu2025physics} and the algorithmic task benchmark CLRS-Text~\cite{deepmind2024clrstext}. Next, we briefly introduce the synthetic algorithmic tasks.

\textbf{MANO:} Each instance of \textsc{MANO} is a randomly generated modular-arithmetic expression in prefix notation with operators $\{+,-,\times\}$ and operands in $\{0,\dots,22\}$. The target is the expression value modulo $23$. A prefix expression with $L$ binary operators can be evaluated in a single left-to-right pass that performs exactly $L$ reductions, so the required computation is $\Theta(L)$. Thus, the \textit{complexity knob} is $L$.

\textbf{BREVO:} Each instance of \textsc{BREVO} is a directed acyclic graph (DAG) together with a query node. The target is the set of nodes that the query \emph{depends on}, listed in a specific topological order.\footnote{Predecessors are traversed in the lexicographical order by node name, and the query node itself is excluded from the output.} Computing the transitive predecessor set of the query can be done by a graph traversal over the reachable subgraph, with a time  complexity of $\Theta(|V_{\mathrm{reach}}|+|E_{\mathrm{reach}}|)$. Under the benchmark generator, in-/out-degrees are bounded to \ $\le4$, so the expected traversal cost scales approximately linearly with $N$. Thus the number of nodes $N$ in the DAG serves as the \textit{complexity knob}.

\textbf{CLRS:} The CLRS-Text dataset provides a suite of 30 algorithmic tasks with systematic variation in input size. The training set sparsely covers the size range, enabling analysis of both interpolation and extrapolation. The dataset also provides algorithm traces that provide step-by-step supervision (akin to chain-of-thought). As we focus on latent reasoning, we remove algorithm traces and train from questions alone. This merges trace-distinct variants (e.g., quicksort and bubble sort into a single \textsc{sort} task), yielding 23 unique tasks. The \textit{complexity knob} of each task in CLRS is the size of the input. For example, the complexity knob of the \textsc{SORT} task is defined as the number of items to sort. A full list of problems and the definition of their input size is provided in the Appendix~\ref{app:clrs_knobs}.

\subsection{Synthetic Language}
\label{sec:exp_lano}
We use the synthetic language task \textsc{LANO} from~\citet{allen-zhu2025physics}, where sequences are generated from a probabilistic context-free grammar (PCFG). Unlike the algorithmic tasks, \textsc{LANO} has no single difficulty knob; difficulty varies across token positions as the prefix becomes more or less syntactically ambiguous.

We therefore construct token-level difficulty proxies from an incremental parser. Specifically, we run an incremental prefix parser for PCFGs~\cite{nowak-cotterell-2023-fast} on model-generated strings and record per-token signals. We group these into (i) \emph{ambiguity proxies} capturing parse-space branching and consolidation (\emph{parse-space expansion} and \emph{parse convergence}) and (ii) \emph{compute proxies} measuring parser work (the number of \emph{additions} and \emph{multiplications} per token). Ideally, compute allocation should correlate positively with \emph{parse-space expansion} and compute proxies, and negatively with \emph{parse convergence}. Formal definitions are provided in Appendix~\ref{app:parser-features}.

\subsection{Defining Compute Allocation}
For \textsc{MANO}, \textsc{BREVO} and \textsc{CLRS}, we define compute allocation as the mean
allocated depth $\bar d$ on the answer tokens
using the inference-time depth selection criteria described in Section~\ref{sec:training_depth_sampling}. 
For \textsc{LANO}, we define it as the expected depth $\mathbb{E}_{q_i}[d]$ for each token position.\footnote{We use expected depth here to reduce the variance from depth sampling as we only have proxies for the complexity.}

\begin{figure*}[t]
  \centering
  \begin{subfigure}[t]{0.24\textwidth}
    \centering
    \includegraphics[width=\linewidth]{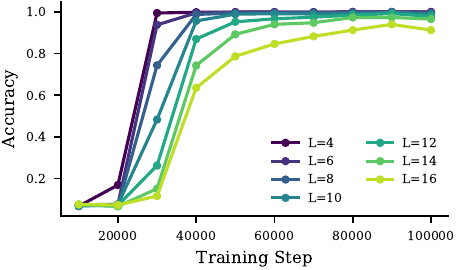}
    \caption{ANIRA-E: Accuracy vs Training steps}
  \end{subfigure}\hfill
  \begin{subfigure}[t]{0.24\textwidth}
    \centering
    \includegraphics[width=\linewidth]{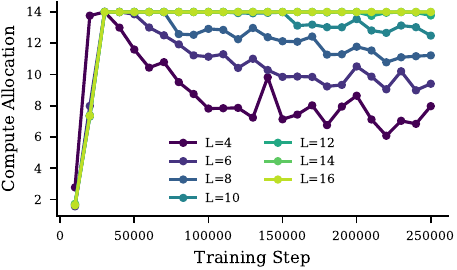}
    \caption{ANIRA-E: Compute Allocation vs Training steps}
  \end{subfigure}\hfill
  \begin{subfigure}[t]{0.24\textwidth}
    \centering
    \includegraphics[width=\linewidth]{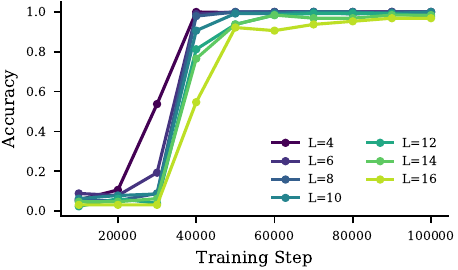}
    \caption{ANIRA-O: Accuracy vs Training steps}
  \end{subfigure}\hfill
  \begin{subfigure}[t]{0.24\textwidth}
    \centering
    \includegraphics[width=\linewidth]{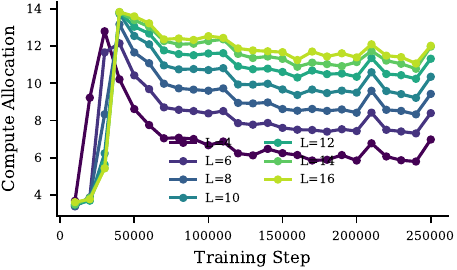}
    \caption{ANIRA-O: Compute Allocation vs Training steps}
  \end{subfigure}

  \caption{MANO Training Dynamics: ANIRA learn tasks in easy-to-hard order and learning happens in two phases: \textit{learning} and \textit{compute reduction}.}
  \label{fig:training_dynamics_mano_row}
\end{figure*}

\begin{figure*}[t]
  \centering
  \begin{subfigure}[t]{0.24\textwidth}
    \centering
    \includegraphics[width=\linewidth]{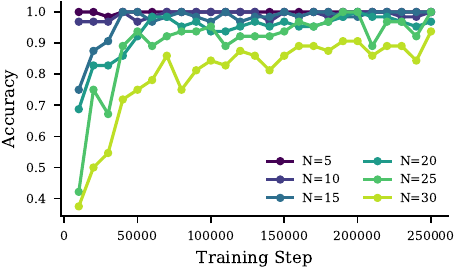}
    \caption{ANIRA-E: Accuracy vs Training steps}
  \end{subfigure}\hfill
  \begin{subfigure}[t]{0.24\textwidth}
    \centering
    \includegraphics[width=\linewidth]{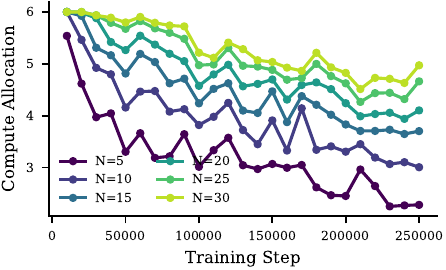}
    \caption{ANIRA-E: Compute Allocation vs Training steps}
  \end{subfigure}\hfill
  \begin{subfigure}[t]{0.24\textwidth}
    \centering
    \includegraphics[width=\linewidth]{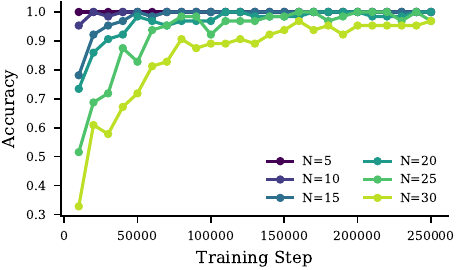}
    \caption{ANIRA-O: Accuracy vs Training steps}
  \end{subfigure}\hfill
  \begin{subfigure}[t]{0.24\textwidth}
    \centering
    \includegraphics[width=\linewidth]{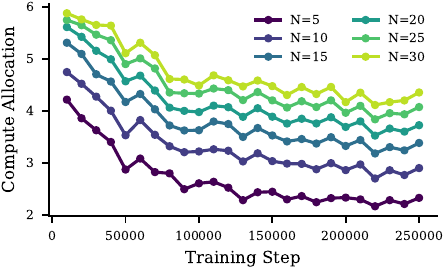}
    \caption{ANIRA-O: Compute Allocation vs Training steps}
  \end{subfigure}

  \caption{BREVO Training Dynamics: ANIRA learns tasks in easy-to-hard order and learning happens in two phases: \textit{learning} and \textit{compute reduction}.}
  \label{fig:training_dynamics_brevo_row}
\end{figure*}
\section{Results}
\subsection{Compute Allocation Tracks Task Complexity}
\label{sec:exp_compute_tracks_complexity}

We train the ANIRA models on MANO instances spanning difficulty levels
$L\in\{3,\dots,16\}$ and separately on BREVO instances with sizes
$N\in\{3,\dots,30\}$. On \textsc{CLRS}, we train \textit{multitask} ANIRA models across all problems in the CLRS training dataset. On \textsc{LANO}, we train ANIRA models on the PCFG 3b\footnote{PCFG 3b has 16 nonterminals, 3 terminals symbols, and 35 production rules with a maximum derivation depth of 7. All branching rules have uniform probability 0.5.} described in~\citet{allen-zhu2025physics}. We set the maximum number of recurrent iterations to $D=6$ for all tasks, except \textsc{MANO} where we use $D=14$. \footnote{For more details see Table~\ref{tab:exp_model_configs} in the Appendix.}

Figure~\ref{fig:compute_allocation_comparison} shows that the ANIRA models learn to allocate compute consistent with task complexity on the \textsc{MANO} and the \textsc{BREVO} tasks.\footnote{Appendix~\ref{app:exp_depo} provides corroborating results on \textsc{DEPO} task.}  Figure~\ref{fig:clrs_selected} shows the same for a representative selection of tasks from the \textsc{CLRS} dataset.\footnote{Appendix ~\ref{app:clrs_knobs} provides results on all CLRS tasks.}  Table~\ref{tab:lano_spearman_expected_correlations} 
shows that the ANIRA models' compute allocation is consistent with the token level prediction complexity on synthetic language. Thus we can conclude that the \textit{ANIRA models allocate compute consistent with complexity even in the absence of explicit complexity signals during training.} 

\subsection{Adaptivity does not lead to algorithmic generalization}
\label{sec:generalization}
As compute allocation tracks task complexity, we might expect ANIRA to learn input-size-invariant algorithmic representations and thus generalize better.
However, the per-size accuracy curves in Figure~\ref{fig:clrs_selected} suggest otherwise: both variants exhibit sharp accuracy drops at unseen input sizes, even within the interpolation regime. This indicates that the \textit{models learn size-specific solutions rather than fully exploiting shared algorithmic structure}.

\subsection{Training Dynamics}
\label{sec:exp_depth_dynamics}
To understand how ANIRA learns its compute policy, we study the training dynamics of task accuracy and depth allocation on the \textsc{MANO} and \textsc{BREVO} tasks. It is interesting to observe from Figure~\ref{fig:training_dynamics_mano_row} and Figure~\ref{fig:training_dynamics_brevo_row} that \textit{the models learn the tasks in an easy-to-hard order}, without explicit supervision.

Further, we see two consistent phases of training: \textit{learning} and \textit{compute reduction}. In the \textit{learning} phase, both variants rapidly increase their compute allocation, often approaching the maximum, suggesting that \textit{the model initially relies on near-full recurrent computation to reduce the task loss}. In the \textit{compute reduction} phase, \textit{the models learn to reduce compute allocation while preserving task performance}. 

\subsection{ANIRA-E and ANIRA-O learn different compute allocation policies}

\label{sec:brevo_answer_compute_analysis}
\textit{For many tasks, especially when trained per-task (e.g., MANO, BREVO, DEPO) ANIRA-O tends to converge to lower average depth than ANIRA-E, which is consistent with online halting leveraging intermediate recurrent states to stop more aggressively}. To further understand the difference in compute allocation policies of ANIRA-E and ANIRA-O, we perform regression analysis on the compute allocation of answer tokens on the \textsc{BREVO} task, against structural and algorithmic state features, extracting features at each token position including hub structure (out-degree), DFS traversal depth, search frontier size and the number of newly enabled nodes. 

Table~\ref{tab:brevo_coefficients} shows the linear regression coefficients of each feature controlling for graph size.
\textit{Despite similar overall performance, the two models employ fundamentally different strategies.} ANIRA-E's compute allocation is primarily explained by the structural feature hub structure, with algorithmic state features contributing minimally. In contrast, ANIRA-O compute allocation is explained primarily by algorithmic execution features DFS depth, frontier size, and newly enabled nodes suggesting it learns to \emph{track algorithmic progress to decide when to halt}.

The architectural choice thus induces qualitatively different learned behaviors. ANIRA-E exploits statistical correlations between structure and difficulty, while ANIRA-O's strategy aligns more closely with the underlying algorithm. Appendix~\ref{app:mechanistic_details} describes the complete methodology and feature definitions.

\subsection{ANIRA starts solving the problem while processing the question tokens}

\begin{table}[t]
\centering
\caption{\textbf{BREVO: answer-token feature coefficients for compute allocation.}
Linear regression coefficients predicting per-answer-token allocated compute from BREVO features, controlling for graph size $N$, larger absolute coefficients indicate stronger association. ANIRA-E aligns most with hub out-degree, while ANIRA-O aligns more with algorithmic-state features. Fit: ANIRA-E $R^2{=}0.7337$, ANIRA-O $R^2{=}0.7088$.}
\label{tab:brevo_coefficients}
\begin{tabular}{lrr}
\toprule
Feature & ANIRA-E & ANIRA-O \\
\midrule
Hub structure (out-degree) & +0.134 & +0.168 \\
DFS traversal depth & $-$0.005 & +0.302 \\
Search frontier size & +0.047 & +0.171 \\
Newly enabled nodes & $-$0.079 & $-$0.286 \\
\bottomrule
\end{tabular}
\end{table}
\label{sec:mano_online_compute}

For the MANO prefix-notation arithmetic evaluation task, the task could be solved while processing the question tokens.
The adaptive nature of the ANIRA models naturally allows us to study whether the models actually learn this.

We derive \emph{prefix-observable} parse-state features from the question prefix using a deterministic stack parser.
For each question token $t$, we record:
(i) \emph{pre-token operator stack depth} $d_t$,
(ii) \emph{remaining operands for current operator} $r_t\in\{0,1,2\}$, the number of operands still required by the current top-of-stack operator immediately before consuming token $t$ (setting $r_t=0$ for the root operator token), and
(iii) \emph{pre-token completed subtree size} $s_t$, defined as the total number of operator nodes in all operator-subtrees completed at the preceding token.

\begin{figure}[t]
  \centering
  \includegraphics[width=0.49\linewidth]{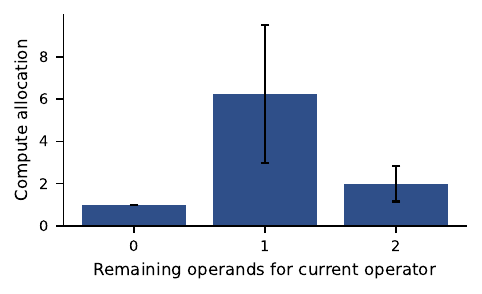}
  \includegraphics[width=0.49\linewidth]{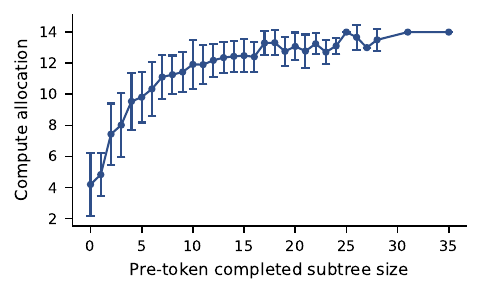}
  \caption{\textbf{MANO question tokens: ANIRA-E compute allocation tracks online parse state.}
  \textbf{Left:} Operator-token compute allocation vs.\ \emph{remaining operands for current operator} $r_t$ before the token ($r_t=0$ root operator token; $r_t=2$ left-child context; $r_t=1$ right-child context).
  \textbf{Right:} For operators with $r_t=1$, compute allocation increases with \emph{pre-token completed subtree size} $s_t$.
  Error bars denote standard deviation; results use only correctly answered instances.}
  \label{fig:mano_question_compute_parse_state}
\end{figure}

Figure~\ref{fig:mano_question_compute_parse_state} shows that ANIRA-E compute allocation is strongly structured by these online state variables: operator tokens receive the most compute when $r_t=1$, and within this subset, compute increases with the size of the immediately preceding completion event $s_t$. This suggests that the model allocates extra compute \emph{while processing the question tokens} as the parse state evolves, rather than deferring all computation to the final answer token.

\subsection{Natural language mathematical questions}
\label{sec:gsm_symbolic}

We study adaptive depth allocation in a natural-language setting by augmenting an eight recurrent layer depth-recurrent Llama model\footnote{\url{https://huggingface.co/smcleish/Recurrent-Llama-3.2-train-recurrence-8}} from~\cite{mcleish2025teachingpretrainedlanguagemodels} with both ANIRA-E and ANIRA-O. We finetune each model for about 5B tokens on Nemotron-CC-Math-v1-4plus~\cite{mahabadi2025nemotronccmath133billiontokenscalehigh}, a large-scale high-quality math corpus. We evaluate on GSM-Symbolic~\cite{mirzadeh2025gsmsymbolic}, a dataset of grade-school-level mathematical questions created from symbolic templates. Relevant to our study, GSM-Symbolic provides three evaluation splits, \textsc{Main}, \textsc{P1}, and
\textsc{P2}, where P1 and P2 add one and two additional clauses to a question relative to \textsc{Main}. We treat these splits as increasing in difficulty,
\textsc{Main} $<$ \textsc{P1} $<$ \textsc{P2}. 

\begin{table}[t]
\centering
\caption{GSM-Symbolic evaluation after finetuning on Nemotron-CC-Math-v1-4plus. We report exact accuracy and the mean allocated depth $\bar d$ over answer tokens. The baseline uses fixed depth $D{=}8$, while ANIRA-E and ANIRA-O allocate variable depth. P1 and P2 add one and two additional clauses relative to \textsc{Main}.}
\small
\setlength{\tabcolsep}{5pt}
\begin{tabular}{l c cc cc}
\toprule
& Baseline & \multicolumn{2}{c}{ANIRA-E} & \multicolumn{2}{c}{ANIRA-O} \\
\cmidrule(lr){2-2}\cmidrule(lr){3-4}\cmidrule(lr){5-6}
Split & Acc.\ (\%) & Acc.\ (\%) & $\bar d$ & Acc.\ (\%) & $\bar d$ \\
\midrule
Main & 46.26 & 45.70 & 6.060 & 43.68 & 4.108 \\
P1   & 23.54 & 24.06 & 6.004 & 23.02 & 4.105 \\
P2   &  5.80 &  5.80 & 5.982 &  5.08 & 4.001 \\
\bottomrule
\end{tabular}
\label{tab:gsm_symbolic_splits}
\end{table}

Table~\ref{tab:gsm_symbolic_splits}
reports exact accuracy and mean allocated depth over answer tokens. Neither ANIRA variant allocates more depth on the harder splits. At the same time, both variants incur only a modest accuracy decrease relative to the depth-$8$ baseline on each split, while achieving substantial reductions in compute allocation. 

We note that measuring task difficulty is inherently complex for natural language tasks. This experiment shows that compute allocation by ANIRA is not easily interpretable on natural language, and does not appear to correlate with a simple notion of task complexity.

\section{Conclusion}

Token-level adaptive computation has largely been treated as an efficiency mechanism evaluated through task-level performance, despite making claims about token-level behavior. This work reframes adaptive computation as a measurement and interpretability problem, arguing that its validity depends on the availability of well-defined, testable notions of token-level difficulty. Our findings suggest that while adaptive compute policies can be learned, their meaning and generalizability are tightly coupled to the structure of the evaluation setting: where token-level complexity is explicit, allocation reflects meaningful signals; where it is not, such as in natural language, interpretation becomes inherently ambiguous. More broadly, this highlights the need for complexity-aware evaluation protocols, principled difficulty proxies, and inductive biases that connect computation to algorithmic structure, if adaptive computation is to serve as a foundation for reliable and interpretable reasoning systems rather than a heuristic optimization.

\section*{Impact Statement}
This paper presents work whose goal is to advance the field of machine learning, specifically by studying token-level adaptive computation for language models and proposing a complexity-controlled evaluation protocol. Our methods may help reduce inference cost and energy use by allocating less computation to easier tokens. The broader societal implications are similar to those of prior work on more efficient model architectures and inference systems, and we do not anticipate impacts beyond these well-established considerations.

\bibliography{example_paper}

@inproceedings{nowak-cotterell-2023-fast,
    title = "A Fast Algorithm for Computing Prefix Probabilities",
    author = "Nowak, Franz  and
      Cotterell, Ryan",
    editor = "Rogers, Anna  and
      Boyd-Graber, Jordan  and
      Okazaki, Naoaki",
    booktitle = "Proceedings of the 61st Annual Meeting of the Association for Computational Linguistics (Volume 2: Short Papers)",
    month = jul,
    year = "2023",
    address = "Toronto, Canada",
    publisher = "Association for Computational Linguistics",
    url = "https://aclanthology.org/2023.acl-short.6/",
    doi = "10.18653/v1/2023.acl-short.6",
    pages = "57--69"
}

@inproceedings{
geiping2025scaling,
title={Scaling up Test-Time Compute with Latent Reasoning: A Recurrent Depth Approach},
author={Jonas Geiping and Sean Michael McLeish and Neel Jain and John Kirchenbauer and Siddharth Singh and Brian R. Bartoldson and Bhavya Kailkhura and Abhinav Bhatele and Tom Goldstein},
booktitle={The Thirty-ninth Annual Conference on Neural Information Processing Systems},
year={2025},
url={https://openreview.net/forum?id=S3GhJooWIC}
}

@inproceedings{
jang2017categorical,
title={Categorical Reparameterization with Gumbel-Softmax},
author={Eric Jang and Shixiang Gu and Ben Poole},
booktitle={International Conference on Learning Representations},
year={2017},
url={https://openreview.net/forum?id=rkE3y85ee}
}

@inproceedings{
bae2025mixtureofrecursions,
title={Mixture-of-Recursions: Learning Dynamic Recursive Depths for Adaptive Token-Level Computation},
author={Sangmin Bae and Yujin Kim and Reza Bayat and Sungnyun Kim and Jiyoun Ha and Tal Schuster and Adam Fisch and Hrayr Harutyunyan and Ziwei Ji and Aaron Courville and Se-Young Yun},
booktitle={The Thirty-ninth Annual Conference on Neural Information Processing Systems},
year={2025},
url={https://openreview.net/forum?id=QuqsEIVWIG}
}

@article{raposo2024mixture,
  title={Mixture-of-depths: Dynamically allocating compute in transformer-based language models},
  author={Raposo, David and Ritter, Sam and Richards, Blake and Lillicrap, Timothy and Humphreys, Peter Conway and Santoro, Adam},
  journal={arXiv preprint arXiv:2404.02258},
  year={2024}
}

@inproceedings{
Elbayad2020Depth-Adaptive,
title={Depth-Adaptive Transformer},
author={Maha Elbayad and Jiatao Gu and Edouard Grave and Michael Auli},
booktitle={International Conference on Learning Representations},
year={2020},
url={https://openreview.net/forum?id=SJg7KhVKPH}
}

@article{graves2016adaptive,
  title={Adaptive computation time for recurrent neural networks},
  author={Graves, Alex},
  journal={arXiv preprint arXiv:1603.08983},
  year={2016}
}

@inproceedings{
banino2021pondernet,
title={PonderNet: Learning to Ponder},
author={Andrea Banino and Jan Balaguer and Charles Blundell},
booktitle={8th ICML Workshop on Automated Machine Learning (AutoML) },
year={2021},
url={https://openreview.net/forum?id=1EuxRTe0WN}
}

@article{zhu2025scaling,
  title={Scaling latent reasoning via looped language models},
  author={Zhu, Rui-Jie and Wang, Zixuan and Hua, Kai and Zhang, Tianyu and Li, Ziniu and Que, Haoran and Wei, Boyi and Wen, Zixin and Yin, Fan and Xing, He and others},
  journal={arXiv preprint arXiv:2510.25741},
  year={2025}
}

@article{schuster2022confident,
  title={Confident adaptive language modeling},
  author={Schuster, Tal and Fisch, Adam and Gupta, Jai and Dehghani, Mostafa and Bahri, Dara and Tran, Vinh and Tay, Yi and Metzler, Donald},
  journal={Advances in Neural Information Processing Systems},
  volume={35},
  pages={17456--17472},
  year={2022}
}

@inproceedings{dehghani2019universal,
  title={Universal Transformers},
  author={Dehghani, Mostafa and Gouws, Stephan and Vinyals, Oriol and Uszkoreit, Jakob and Kaiser, {\L}ukasz},
  booktitle={International Conference on Learning Representations},
  year={2019},
  url={https://openreview.net/pdf?id=HyxXZwJcwH}
}

@inproceedings{
allen-zhu2025physics,
title={Physics of Language Models: Part 4.1, Architecture Design and the Magic of Canon Layers},
author={Zeyuan Allen-Zhu},
booktitle={The Thirty-ninth Annual Conference on Neural Information Processing Systems},
year={2025},
url={https://openreview.net/forum?id=kxv0M6I7Ud}
}

@article{deepmind2024clrstext,
  title={The CLRS-Text Algorithmic Reasoning Language Benchmark},
  author={Larisa Markeeva and Sean McLeish and Borja Ibarz and Wilfried Bounsi
    and Olga Kozlova and Alex Vitvitskyi and Charles Blundell and
    Tom Goldstein and Avi Schwarzschild and Petar Veli\v{c}kovi\'{c}},
  journal={arXiv preprint arXiv:2406.04229},
  year={2024}
}

@inproceedings{
wei2022chain,
title={Chain of Thought Prompting Elicits Reasoning in Large Language Models},
author={Jason Wei and Xuezhi Wang and Dale Schuurmans and Maarten Bosma and brian ichter and Fei Xia and Ed H. Chi and Quoc V Le and Denny Zhou},
booktitle={Advances in Neural Information Processing Systems},
editor={Alice H. Oh and Alekh Agarwal and Danielle Belgrave and Kyunghyun Cho},
year={2022},
url={https://openreview.net/forum?id=_VjQlMeSB_J}
}

@misc{hao2025traininglargelanguagemodels,
      title={Training Large Language Models to Reason in a Continuous Latent Space}, 
      author={Shibo Hao and Sainbayar Sukhbaatar and DiJia Su and Xian Li and Zhiting Hu and Jason Weston and Yuandong Tian},
      year={2025},
      eprint={2412.06769},
      archivePrefix={arXiv},
      primaryClass={cs.CL},
      url={https://arxiv.org/abs/2412.06769}, 
}

@misc{mcleish2025teachingpretrainedlanguagemodels,
      title={Teaching Pretrained Language Models to Think Deeper with Retrofitted Recurrence}, 
      author={Sean McLeish and Ang Li and John Kirchenbauer and Dayal Singh Kalra and Brian R. Bartoldson and Bhavya Kailkhura and Avi Schwarzschild and Jonas Geiping and Tom Goldstein and Micah Goldblum},
      year={2025},
      eprint={2511.07384},
      archivePrefix={arXiv},
      primaryClass={cs.CL},
      url={https://arxiv.org/abs/2511.07384}, 
}

@misc{mahabadi2025nemotronccmath133billiontokenscalehigh,
      title={Nemotron-CC-Math: A 133 Billion-Token-Scale High Quality Math Pretraining Dataset}, 
      author={Rabeeh Karimi Mahabadi and Sanjeev Satheesh and Shrimai Prabhumoye and Mostofa Patwary and Mohammad Shoeybi and Bryan Catanzaro},
      year={2025},
      eprint={2508.15096},
      archivePrefix={arXiv},
      primaryClass={cs.CL},
      url={https://arxiv.org/abs/2508.15096}, 
}

@inproceedings{
mirzadeh2025gsmsymbolic,
title={{GSM}-Symbolic: Understanding the Limitations of Mathematical Reasoning in Large Language Models},
author={Seyed Iman Mirzadeh and Keivan Alizadeh and Hooman Shahrokhi and Oncel Tuzel and Samy Bengio and Mehrdad Farajtabar},
booktitle={The Thirteenth International Conference on Learning Representations},
year={2025},
url={https://openreview.net/forum?id=AjXkRZIvjB}
}

@article{vaswani2017attention,
  title={Attention is all you need},
  author={Vaswani, Ashish and Shazeer, Noam and Parmar, Niki and Uszkoreit, Jakob and Jones, Llion and Gomez, Aidan N and Kaiser, {\L}ukasz and Polosukhin, Illia},
  journal={Advances in neural information processing systems},
  volume={30},
  year={2017}
}

@inproceedings{
loshchilov2018decoupled,
title={Decoupled Weight Decay Regularization},
author={Ilya Loshchilov and Frank Hutter},
booktitle={International Conference on Learning Representations},
year={2019},
url={https://openreview.net/forum?id=Bkg6RiCqY7},
}

@misc{bengio2013estimatingpropagatinggradientsstochastic,
      title={Estimating or Propagating Gradients Through Stochastic Neurons for Conditional Computation}, 
      author={Yoshua Bengio and Nicholas Léonard and Aaron Courville},
      year={2013},
      eprint={1308.3432},
      archivePrefix={arXiv},
      primaryClass={cs.LG},
      url={https://arxiv.org/abs/1308.3432}, 
}
\bibliographystyle{icml2026}

\newpage
\appendix
\onecolumn
\section{Appendix}
\subsection{CLRS-text Task Complexity Knobs}
\label{app:clrs_knobs}
\begin{table*}[t]
\centering
\small
\begin{tabular}{l p{0.72\linewidth}}
\hline
Problem & Input size $n$ (definition) \\
\hline
Sort & $n = |\mathbf{x}|$ for an input array $\mathbf{x}=(x_1,\dots,x_n)$. \\
Binary search & $n = |\mathbf{x}|$ for a sorted array $\mathbf{x}=(x_1,\dots,x_n)$. \\
Find minimum & $n = |\mathbf{x}|$ for an input array $\mathbf{x}=(x_1,\dots,x_n)$. \\
Select (order statistic) & $n = |\mathbf{x}|$ for an input array $\mathbf{x}=(x_1,\dots,x_n)$. \\
Maximum subarray & $n = |\mathbf{x}|$ for an input array $\mathbf{x}=(x_1,\dots,x_n)$. \\
String matching & $n = |T|=|P|$ for text $T\in\Sigma^{n}$ and pattern $P\in\Sigma^{n}$. \\
Longest common subsequence & $n = |X|=|Y|$ for sequences $X,Y\in\Sigma^{n}$. \\
Matrix chain multiplication & $n = |p|$ for dimension vector $p=(p_0,\dots,p_{n-1})$ (so \#matrices $=n-1$). \\
Optimal BST & $n$ = number of keys, with probabilities $\{p_i\}_{i=1}^{n}$ and $\{q_i\}_{i=0}^{n}$. \\
Activity selection & $n$ = number of activities/intervals $\{(s_i,f_i)\}_{i=1}^{n}$. \\
Task scheduling & $n$ = number of jobs $\{(d_i,w_i)\}_{i=1}^{n}$. \\
Convex hull & $n$ = number of planar points $\{(x_i,y_i)\}_{i=1}^{n}\subset\mathbb{R}^2$. \\
Shortest path & $n = |V|$ for a graph $G=(V,E)$ (vertices). \\
MST Kruskal & $n = |V|$ for a graph $G=(V,E)$ (vertices); outputs edge matrix. \\
MST Prim & $n = |V|$ for a graph $G=(V,E)$ (vertices); outputs predecessor array. \\
Breadth-first search & $n = |V|$ for a graph $G=(V,E)$ (vertices). \\
Depth-first search & $n = |V|$ for a graph $G=(V,E)$ (vertices). \\
Topological sort & $n = |V|$ for a directed graph $G=(V,E)$ (vertices). \\
Strongly connected components & $n = |V|$ for a directed graph $G=(V,E)$ (vertices). \\
All-pairs shortest paths & $n = |V|$ for a graph $G=(V,E)$ (vertices). \\
Articulation points & $n = |V|$ for a graph $G=(V,E)$ (vertices). \\
Bridges & $n = |V|$ for a graph $G=(V,E)$ (vertices). \\
Bipartite matching & $n = |U|+|W|$ for bipartite $G=(U,W,E)$. \\
Segment intersection & constant-size instance (two line segments in $\mathbb{R}^2$; no varying $n$). \\
\hline
\end{tabular}
\caption{Definition of input size $n$ per CLRS problem used for our CLRS-Text analysis.}
\label{tab:clrs_input_size_def}
\end{table*}
\begin{figure*}[t]
  \centering
  \begin{subfigure}[t]{0.16\textwidth}
    \centering
    \includegraphics[width=\linewidth]{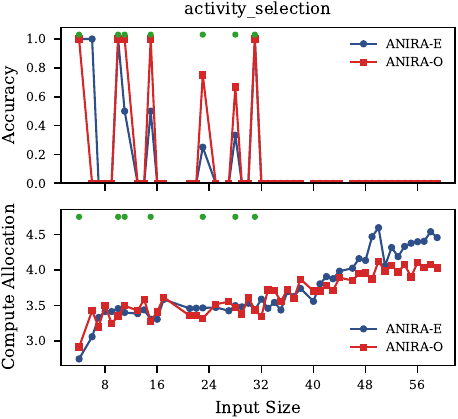}
    \caption{\textsc{activity sel.}}
  \end{subfigure}
  \begin{subfigure}[t]{0.16\textwidth}
    \centering
    \includegraphics[width=\linewidth]{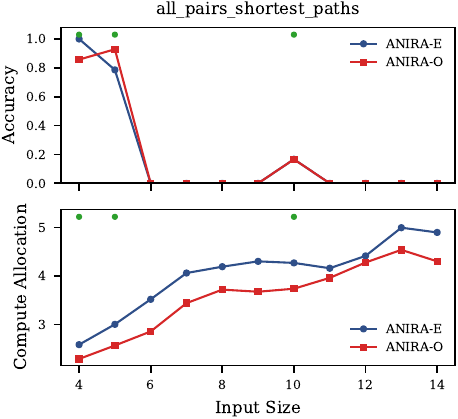}
    \caption{\textsc{apsp}}
  \end{subfigure}
  \begin{subfigure}[t]{0.16\textwidth}
    \centering
    \includegraphics[width=\linewidth]{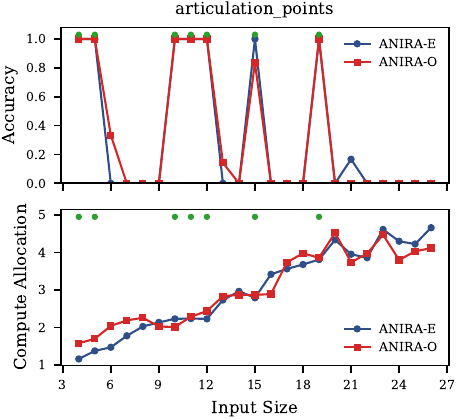}
    \caption{\textsc{artic. points}}
  \end{subfigure}
  \begin{subfigure}[t]{0.16\textwidth}
    \centering
    \includegraphics[width=\linewidth]{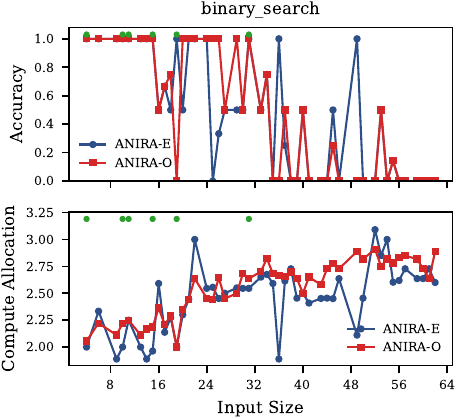}
    \caption{\textsc{binary search}}
  \end{subfigure}
  \begin{subfigure}[t]{0.16\textwidth}
    \centering
    \includegraphics[width=\linewidth]{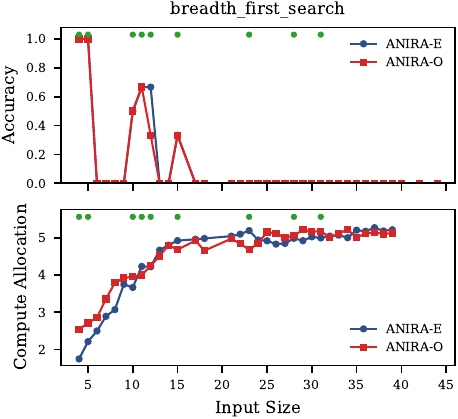}
    \caption{\textsc{bfs}}
  \end{subfigure}
  \begin{subfigure}[t]{0.16\textwidth}
    \centering
    \includegraphics[width=\linewidth]{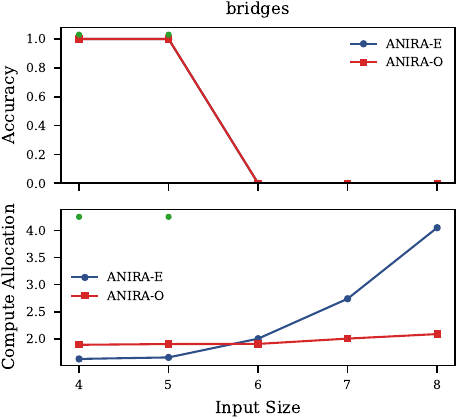}
    \caption{\textsc{bridges}}
  \end{subfigure}
  
  \begin{subfigure}[t]{0.16\textwidth}
    \centering
    \includegraphics[width=\linewidth]{clrs_figures/convex_hull_comparison_subplots.pdf}
    \caption{\textsc{convex hull}}
  \end{subfigure}
  \begin{subfigure}[t]{0.16\textwidth}
    \centering
    \includegraphics[width=\linewidth]{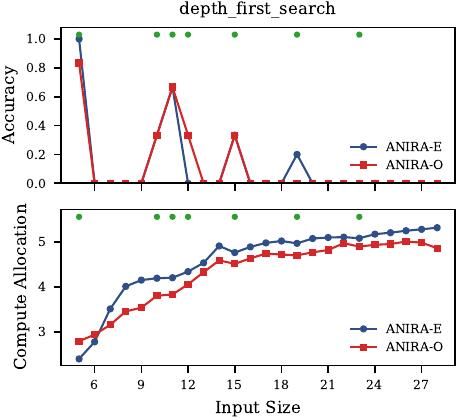}
    \caption{\textsc{dfs}}
  \end{subfigure}
  \begin{subfigure}[t]{0.16\textwidth}
    \centering
    \includegraphics[width=\linewidth]{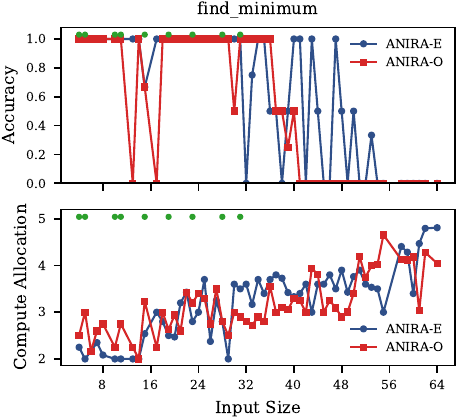}
    \caption{\textsc{find min}}
  \end{subfigure}
  \begin{subfigure}[t]{0.16\textwidth}
    \centering
    \includegraphics[width=\linewidth]{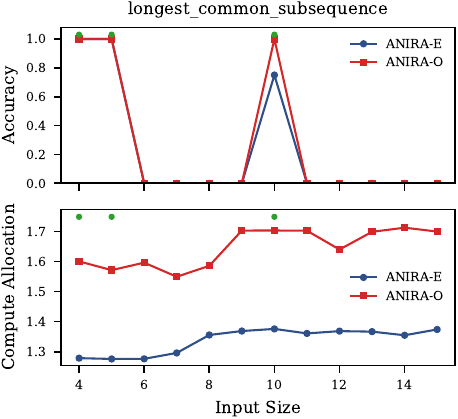}
    \caption{\textsc{lcs}}
  \end{subfigure}
  \begin{subfigure}[t]{0.16\textwidth}
    \centering
    \includegraphics[width=\linewidth]{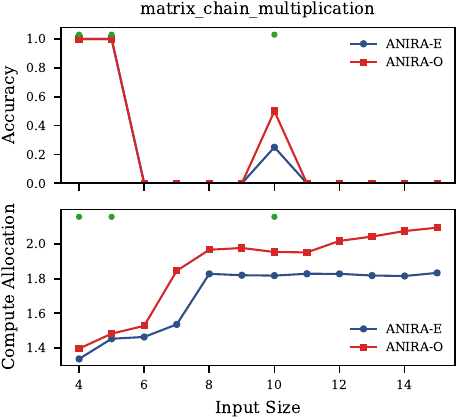}
    \caption{\textsc{matrix chain}}
  \end{subfigure}
  \begin{subfigure}[t]{0.16\textwidth}
    \centering
    \includegraphics[width=\linewidth]{clrs_figures/maximum_subarray_comparison_subplots.pdf}
    \caption{\textsc{max subarray}}
  \end{subfigure}
  
  \begin{subfigure}[t]{0.16\textwidth}
    \centering
    \includegraphics[width=\linewidth]{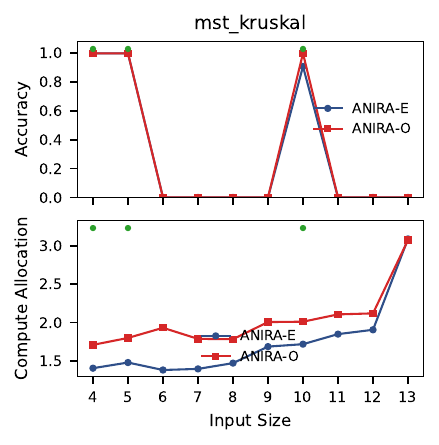}
    \caption{\textsc{mst kruskal}}
  \end{subfigure}
  \begin{subfigure}[t]{0.16\textwidth}
    \centering
    \includegraphics[width=\linewidth]{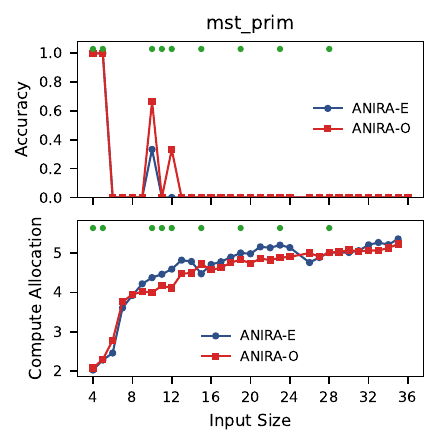}
    \caption{\textsc{mst prim}}
  \end{subfigure}
  \begin{subfigure}[t]{0.16\textwidth}
    \centering
    \includegraphics[width=\linewidth]{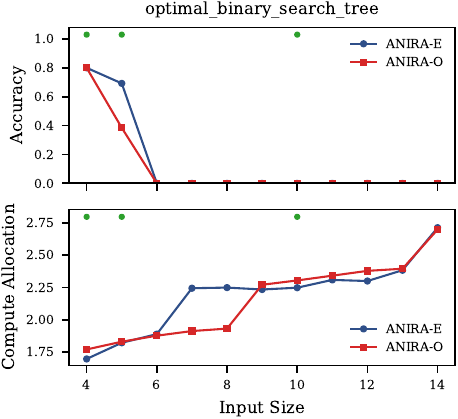}
    \caption{\textsc{opt. bst}}
  \end{subfigure}
  \begin{subfigure}[t]{0.16\textwidth}
    \centering
    \includegraphics[width=\linewidth]{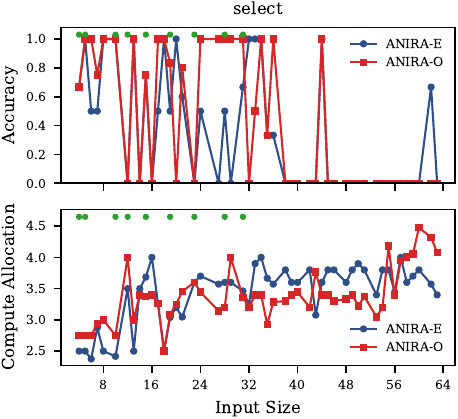}
    \caption{\textsc{select}}
  \end{subfigure}
  \begin{subfigure}[t]{0.16\textwidth}
    \centering
    \includegraphics[width=\linewidth]{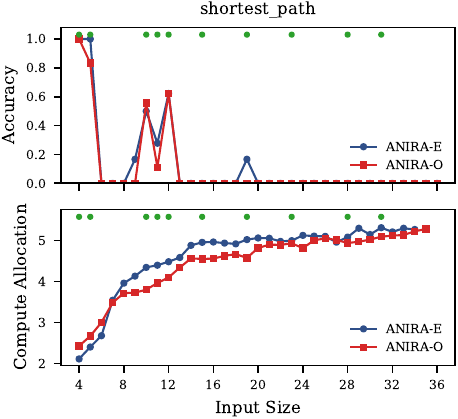}
    \caption{\textsc{shortest path}}
  \end{subfigure}
  \begin{subfigure}[t]{0.16\textwidth}
    \centering
    \includegraphics[width=\linewidth]{clrs_figures/sort_comparison_subplots.pdf}
    \caption{\textsc{sort}}
  \end{subfigure}
  
  \begin{subfigure}[t]{0.16\textwidth}
    \centering
    \includegraphics[width=\linewidth]{clrs_figures/string_matching_comparison_subplots.pdf}
    \caption{\textsc{string match}}
  \end{subfigure}
  \begin{subfigure}[t]{0.16\textwidth}
    \centering
    \includegraphics[width=\linewidth]{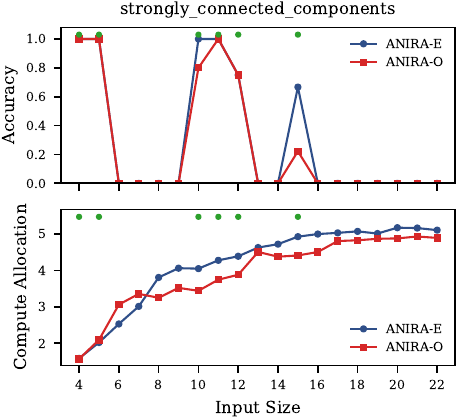}
    \caption{\textsc{scc}}
  \end{subfigure}
  \begin{subfigure}[t]{0.16\textwidth}
    \centering
    \includegraphics[width=\linewidth]{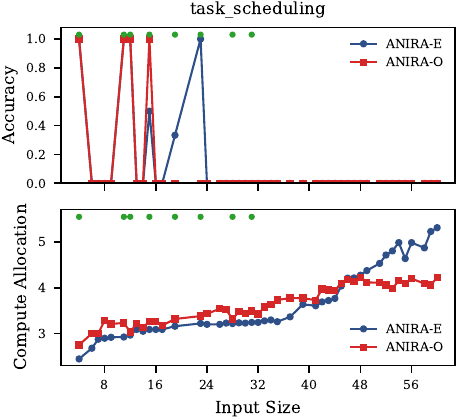}
    \caption{\textsc{task sched.}}
  \end{subfigure}
  \begin{subfigure}[t]{0.16\textwidth}
    \centering
    \includegraphics[width=\linewidth]{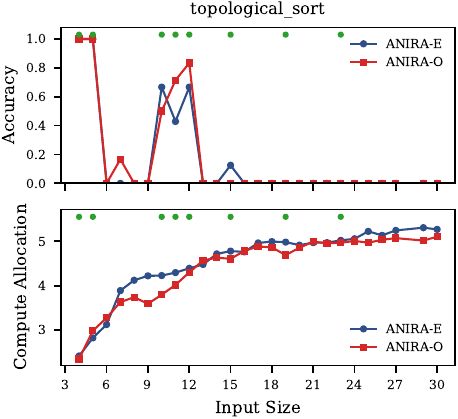}
    \caption{\textsc{topo. sort}}
  \end{subfigure}
  \caption{CLRS task complexity vs accuracy (top) and compute allocation $\bar d$ (bottom) for all 21 tasks. Green markers indicate input sizes seen during training. ANIRA compute allocation tracks task complexity. However, we observe that task accuracy drops sharply at input sizes not covered in training set, indicating interpolation and extrapolation failure. Note, we ignore \textsc{segments intersection} problem as it's complexity does not vary in the dataset.}
  \label{fig:clrs_all}
\end{figure*}

Section~\ref{sec:algorithmic_tasks} presented four representative tasks from the CLRS-Text benchmark. Here we present the results for all algorithmic tasks. Table~\ref{tab:clrs_input_size_def} explains the complexity knob definition for each task of the CLRS-text benchmark. Figure~\ref{fig:clrs_all} shows the accuracy and compute allocation plots for all the algorithmic tasks in the benchmark. Our conclusion from Section~\ref{sec:exp_compute_tracks_complexity} that ANIRA models learn reasonable compute allocation policies holds generally for all CLRS tasks. Also, we see the same generalization failure discussed in Section~\ref{sec:generalization} on all tasks.

\subsection{Model and training configuration}
\begin{table*}[t]
\centering
\small
\begin{tabular}{lcccccccccc}
\toprule
Model & Task & $D$ & \#heads & Prelude/Coda & $d_\text{model}$ & $d_\text{ff}$ & maxlen & batch size & LR & $\gamma/b$ \\
\midrule
\multirow{4}{*}{ANIRA}
& MANO  & 14 & 8 & 1/1 & 256 & 1664 & 64 & 512 & 1e-4 & 1e-1/1.25  \\
& BREVO & 6 & 8 & 1/1 & 512 & 2048 & 256 & 64 & 1e-3 & 1e-1/2.0 \\
& DEPO  & 6 & 8 & 1/1 & 512 & 2048 & 256 & 64 & 3e-4 & 1e-1/2.0 \\
& LANO  & 6 & 8 & 1/1 & 512 & 2048 & 512 & 64 & 1e-4 & 1e-2/1.0 \\
& CLRS  & 6 & 8 & 1/1 & 512 & 2048 & 4096 & 32 & 1e-4 & 1e-2/2.0 \\
\bottomrule
\end{tabular}

\caption{ANIRA hyperparameters. Both ANIRA-E and ANIRA-O employ the same hyperparameters. The hyperparameters LR, $\gamma$ and $b$ were chosen based on the performance on validation sets with a small grid search for ANIRA-E and reused for ANIRA-O.}
\label{tab:exp_model_configs}
\end{table*}

Table~\ref{tab:exp_model_configs} presents the detailed model configuration and hyperparameter used to train the ANIRA models on each task. The ANIRA models were implemented by modifying the Llama 2 model implementation available from Huggingface Transformers library. 

For ANIRA with hidden size 512, each of the Prelude, recurrent core, and Coda has about 4.2M parameters. For MANO we use hidden size 256, where each component has about 1.5M parameters. The corresponding decider module has about 1.0M parameters (hidden size 512) and about 0.4M parameters (hidden size 256), respectively. In total the ANIRA models with hidden size of 512 have about 13.7M parameters and the models with hidden size of 256 have about 5.1M parameters.

The deciders are Feedforward Networks (FFNs) matching the FFNs used elsewhere in the model. The architecture consists of layer normalization with learnable affine parameters, followed by a FFN. The first linear layer projects the H-dimensional hidden state to an intermediate dimension I (set to 4H, except for MANO, for which it is 6.5H) without bias, followed by a SiLU nonlinearity. The second linear layer, which includes a bias term, maps from the intermediate dimension to the output space. The two ANIRA variants differ in their output space. ANIRA-E produces $D$ logits corresponding to a categorical distribution over depth levels. These logits are centered by subtracting their mean before applying the softmax function, which stabilizes training. ANIRA-O instead produces a single logit that is passed through a sigmoid function to obtain the halting probability.

We use the AdamW~\cite{loshchilov2018decoupled} optimizer with a constant learning rate with warmup schedule. Warmup steps were set to 1000. We trained all ANIRA models for 250k steps. Each model was trained on a single NVIDIA A100 GPU. Each training run took about seven hours.

For the natural language experiments in section~\ref{sec:gsm_symbolic}, we train the models for 5B tokens, with a batch size of 32 and sequence length of 4096, learning rate of $2e-5$, $\gamma$ of $1e-1$, on four NVIDIA A100 GPUs, each with 80GB VRAM, which took about a day.

\subsection{Results on DEPO: $K$-Step Successor in a Directed Cyclic Graph}
\label{app:exp_depo}

DEPO evaluates multi-hop retrieval ability.
Each instance defines a directed cycle over $N$ nodes. The
input presents this cycle as a shuffled list of directed edges, followed by up to $\min(N,10)$ queries. Each query specifies a starting node and a step count $K$, and the target is the $K^{\text{th}}$ successor of the start node along the cycle. We serialize queries as \texttt{<query-k> <q> <ans> <y> <eoa>}, where the query token \texttt{<query-k>} encodes $K$ in the token identity (e.g., \texttt{<query-03>}). Node names are single tokens, so each query’s answer is a single token; supervision is applied only to that answer token.

\paragraph{Complexity knob.}
Recovering the $K^{\text{th}}$ successor requires composing $K$ successor steps, yielding an inherent sequential cost $\Theta(K)$.

\begin{figure}
    \centering
    \includegraphics[width=0.5\linewidth]{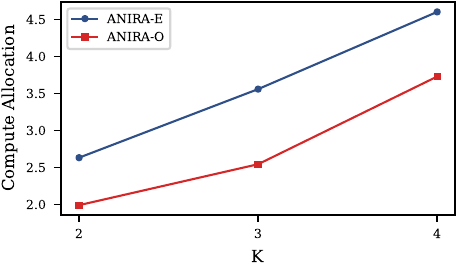}
    \caption{Compute allocation vs. complexity for DEPO}
    \label{fig:depo_compute_allocation_comparison}
\end{figure}

\paragraph{Compute allocation.}
We train a single model on a mixture of instances with varying $N$ and $K$ up to maximum values $50$ and $4$, and then evaluate compute allocation using fixed-$K$ evaluation sets, stratified by $K$ (averaging over the evaluated $N$ values). Using the inference-time depth selections described in Section~\ref{sec:training_depth_sampling}, we compute the mean allocated depth $\bar d$ over the answer tokens. Figure~\ref{fig:depo_compute_allocation_comparison}
show that $\bar d$ increases with $K$ for both variants, indicating that ANIRA learns to allocate more recurrent computation to instances requiring longer iterative retrieval.

\subsection{Extrapolation Beyond the Training Range}
\label{sec:exp_extrapolation_mano}
\begin{figure}
    \centering
    \includegraphics[width=0.5\linewidth]{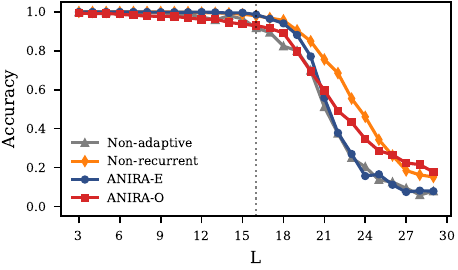}
    \caption{MANO extrapolation: expression length $L$ vs. accuracy. The dashed line
    indicates the maximum $L$ seen during training.}
    \label{fig:mano_extrapolation}
\end{figure}

We probe out-of-distribution generalization by evaluating on MANO instances with expression length $L$ beyond the training range $L\in\{3,\dots,16\}$. Figure~\ref{fig:mano_extrapolation} shows accuracy vs
$L$ for ANIRA-E, ANIRA-O, a non-adaptive model and standard transformer model without weight sharing.

All four models achieve near-perfect accuracy within the training range, but performance degrades sharply once $L$ exceeds the training maximum, indicating limited extrapolation to longer expressions. ANIRA-O degrades more gracefully and maintains higher accuracy than both ANIRA-E and the non-adaptive baseline at the largest $L$, but the overall accuracy remains low in the extrapolation regime. Neither recurrence/weight sharing nor adaptive depth allocation substantially improves extrapolation, although online halting provides a modest robustness benefit.

\subsection{Pareto frontier for ANIRA-O}
\begin{figure}
    \centering
    \includegraphics[width=0.9\linewidth]{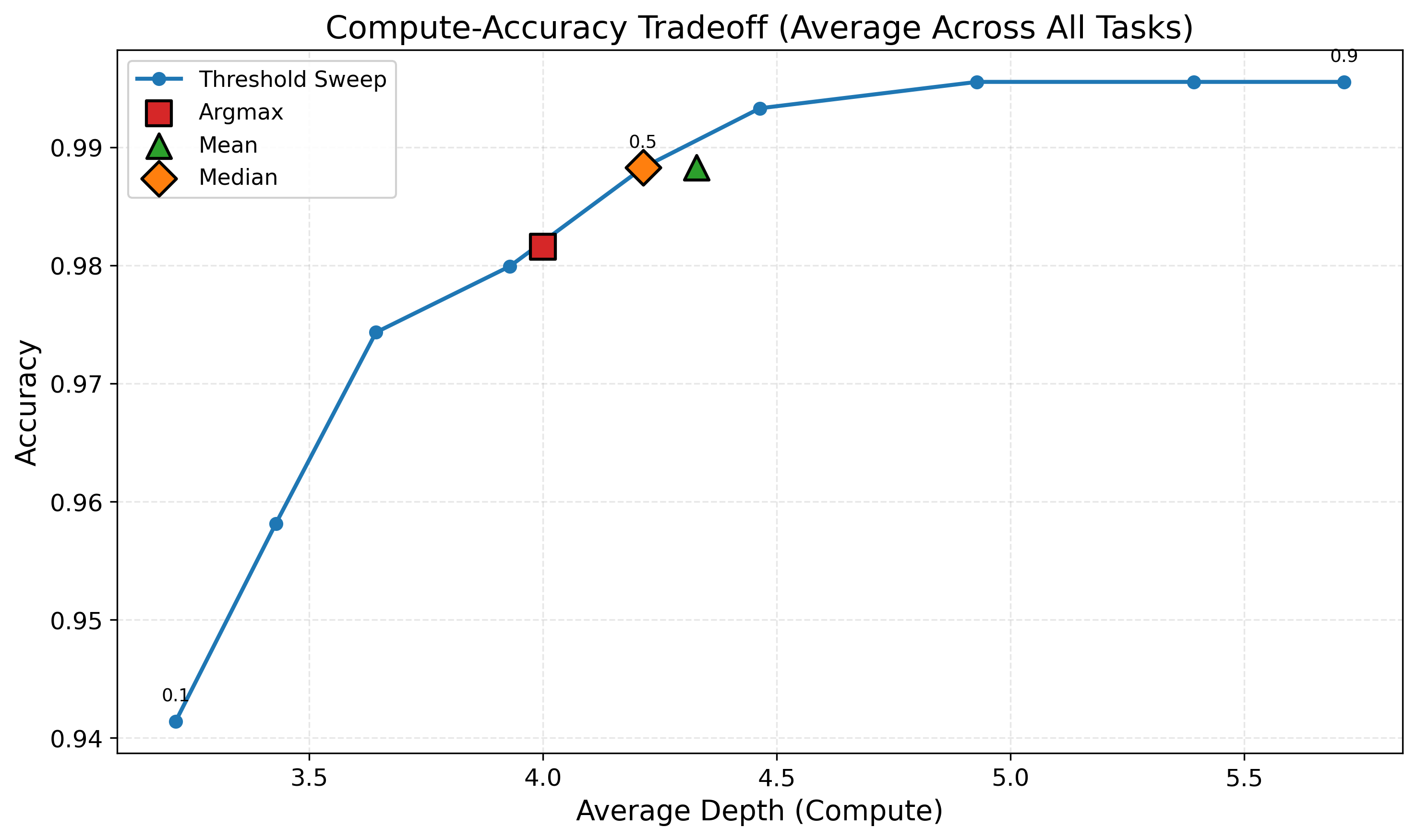}
    \caption{ANIRA-E: Accuracy for different depth choice during inference and cumulative threshold sweep.}
    \label{fig:brevo_threshold_pareto}
\end{figure}

The ANIRA-O model allows  different choices for the depth choice during inference. In the main paper, we exclusively used the mode of the distribution during inference. On Figure~\ref{fig:brevo_threshold_pareto}, we show the model's accuracy for mean and median choice for the BREVO task. The different choices lead to minor variance in accuracy and compute allocation.

Further, we study the compute-accuracy tradeoff by changing the cumulative threshold. Figure~\ref{fig:brevo_threshold_pareto} shows that by choosing different thresholds we can trade off between task accuracy and compute cost.
\subsection{MANO: Question Token Compute Allocation Piecewise model}

 The following piecewise parametric model was found to fit ANIRA-E question-token compute allocation $\hat{c}(t)$ on MANO:
\begin{equation}
\hat c(t)=
\begin{cases}
2.678 & \text{operand},\ r_t=1\\
1.511 & \text{operand},\ r_t=2\\
1.000 & \text{operator},\ r_t=0\\
4.088 + 0.625\,s_t + 0.261\,d_t & \text{operator},\ r_t=1\\
1.104 + 0.337\,d_t & \text{operator},\ r_t=2
\end{cases}
\end{equation}
which attains $R^2\approx 0.63$, indicating that most systematic variation in question-time compute is explained by a small set of prefix-observable parse-state features.

\subsection{Incremental Parser Features}
\label{app:parser-features}

We extract syntactic complexity features using an incremental prefix PCFG parser~\cite{nowak-cotterell-2023-fast}. The parser maintains two chart data structures. The first is $\beta[i, X, k]$, which stores \emph{inside probabilities}, i.e., the probability that nonterminal $X$ derives the substring spanning positions $[i, k)$ according to the grammar. The second is an auxiliary structure $\gamma[i, j, Y, Z]$, which accumulates probabilities over partial derivations where nonterminal $Y$ has been recognized over positions $[i, j)$ and nonterminal $Z$ is expected to continue from position $j$. Intuitively, $\gamma$ tracks intermediate parsing states where the parser has identified a left child constituent and is searching for a matching right child.

After appending a token at position $N$, we compute the following features:

\begin{itemize}
    \item \textbf{active\_gamma\_slice} (parse-space expansion): The number of non-zero entries in $\gamma$ at the most recent split point, defined as $|\{(i, Y, Z) : \gamma[i, N{-}1, Y, Z] \neq 0\}|$. This quantity measures the expansion of the parse space, where higher values indicate a greater number of candidate continuations under consideration.
    
    \item \textbf{active\_beta\_end} (parse convergence): The number of non-zero entries in $\beta$ that terminate at the current position, defined as $|\{(i, X) : \beta[i, X, N] \neq 0\}|$. This quantity counts the number of completed constituent spans at the current position.
    
    \item \textbf{ops\_add} and \textbf{ops\_mul}: The total number of addition and multiplication operations, respectively, performed during the processing of the current token. These quantities measure the arithmetic intensity of each parsing step.
\end{itemize}

\subsection{BREVO Answer Token Compute Allocation Analysis Details}
\label{app:mechanistic_details}

This appendix provides complete methodological details and results for the mechanistic analysis of compute allocation strategies in ANIRA-E and ANIRA-O models on the BREVO task, complementing the summary presented in Section~\ref{sec:brevo_answer_compute_analysis}.

\subsubsection{Task Description and Data Collection}

The BREVO task requires the model to output all topological dependencies of a query node in a directed acyclic graph (DAG) following depth-first search (DFS) postorder traversal. Graph instances range from $N=3$ to $N=30$ nodes. We analyze compute allocation on correctly solved instances only, yielding 210,640 answer tokens from 26,641 samples for ANIRA-E and 14,046 answer tokens from 1,718 samples for ANIRA-O.

We define the target variable as expected depth $\mathbb{E}[\text{depth}] = \sum_{d=1}^{D} d \cdot p_d$ where $p_d$ denotes the model's predicted probability of using depth $d$. This continuous measure proves superior to argmax depth, explaining 11\% additional variance while avoiding discretization artifacts.

\subsubsection{Feature Extraction}

At each answer token position, we extract features characterizing both graph structure and algorithmic execution state. Structural features include: total graph size $N$, out-degree dependents (number of nodes depending on the current node, capturing hub structure), in-degree (number of dependencies), and remaining subtree size. Algorithmic state features include: DFS traversal depth, active frontier size (nodes in search queue), newly enabled nodes (dependencies satisfied by current output), and graph distance to query node.

We additionally tested READ-inspired features (average and maximum expected depth of already-output predecessors) to examine whether models must reach depths where predecessors encoded information. These features contributed negligibly (R$^2 < 0.1\%$) and exhibited near-perfect correlation ($r > 0.998$), indicating READ complexity does not drive compute allocation in this task.

\subsubsection{Regression Methodology}

Our analysis proceeds in four stages. First, we address multicollinearity by removing features with pairwise correlations $|r| > 0.8$ (retaining the feature more correlated with the target) and computing Variance Inflation Factors (VIF), retaining only features with VIF $< 5$. This eliminates \texttt{out\_degree\_with\_query} ($r = 0.93$ with \texttt{out\_degree\_dep}) and \texttt{max\_pred\_exp\_depth} ($r = 0.998$ with \texttt{avg\_pred\_exp\_depth}).

Second, we perform ablation analysis, measuring each feature's marginal contribution as $\Delta R^2 = R^2_{\text{full}} - R^2_{\text{without feature}}$. Third, we select features with $\Delta R^2 > 0.5\%$ for the final model. Fourth, we validate feature importance by refitting with $N$ as a categorical variable (one-hot encoded with the first category as baseline), isolating feature effects within graph size to test whether features capture true algorithmic information or merely proxy for problem scale.

\subsubsection{Results: ANIRA-E Model}

Table~\ref{tab:dd_linear} presents ablation results for the linear $N$ model (R$^2 = 68.66\%$). Graph size dominates the model, contributing $\Delta R^2 = 28.29\%$, with coefficient $+0.059$ indicating approximately 0.06 additional layers per node. Hub structure (\texttt{out\_degree\_dep}) contributes $\Delta R^2 = 4.41\%$ with coefficient $+0.152$. Remaining features (subtree size, frontier size, newly enabled nodes) contribute less than 1\% each.

\begin{table}[h]
\centering
\caption{ANIRA-E feature importance with linear $N$ parameterization. Features selected via ablation analysis with $\Delta R^2 > 0.5\%$ threshold.}
\label{tab:dd_linear}
\begin{tabular}{lrr}
\toprule
Feature & $\Delta R^2$ (\%) & Coefficient \\
\midrule
$N$ (graph size) & 28.29 & +0.059 \\
Hub structure & 4.41 & +0.152 \\
Subtree size & 0.70 & $-$0.022 \\
Frontier size & 0.64 & +0.049 \\
Newly enabled & 0.55 & $-$0.081 \\
\midrule
Intercept & --- & +3.21 \\
\bottomrule
\end{tabular}
\end{table}

Table~\ref{tab:dd_categorical} presents results when $N$ is treated as categorical (R$^2 = 73.37\%$). The substantial improvement ($\Delta R^2 = +4.71\%$) indicates strong non-linear size dependence. Hub structure maintains $\Delta R^2 = 3.51\%$ within graph size (ratio 0.80 relative to linear model), confirming it captures genuine structural information beyond size. Other features show larger decreases (ratios 0.84---1.08), suggesting they partially proxy for graph size.

\begin{table}[h]
\centering
\caption{ANIRA-E feature importance with categorical $N$ fixed effects. Ratio indicates $\Delta R^2_{\text{categorical}} / \Delta R^2_{\text{linear}}$, measuring feature retention when controlling for non-linear size effects.}
\label{tab:dd_categorical}
\begin{tabular}{lrr}
\toprule
Feature & $\Delta R^2$ (\%) & Ratio \\
\midrule
Hub structure & 3.51 & 0.80 \\
Subtree size & 0.75 & 1.08 \\
Frontier size & 0.55 & 0.86 \\
Newly enabled & 0.46 & 0.84 \\
\bottomrule
\end{tabular}
\end{table}

\subsubsection{Results: ANIRA-O Model}

Table~\ref{tab:hd_linear} presents ablation results for the linear $N$ model (R$^2 = 70.00\%$). Unlike ANIRA-E, importance is distributed across features. DFS depth leads with $\Delta R^2 = 5.16\%$ (coefficient $+0.303$), followed by graph size ($\Delta R^2 = 4.83\%$, coefficient $+0.032$), newly enabled nodes ($\Delta R^2 = 3.75\%$), frontier size ($\Delta R^2 = 3.63\%$), and hub structure ($\Delta R^2 = 3.02\%$).

\begin{table}[h]
\centering
\caption{ANIRA-O feature importance with linear $N$ parameterization. Note the balanced importance across algorithmic state features, contrasting with ANIRA-E's size-dominated allocation.}
\label{tab:hd_linear}
\begin{tabular}{lrr}
\toprule
Feature & $\Delta R^2$ (\%) & Coefficient \\
\midrule
DFS depth & 5.16 & +0.303 \\
$N$ (graph size) & 4.83 & +0.032 \\
Newly enabled & 3.75 & $-$0.287 \\
Frontier size & 3.63 & +0.170 \\
Hub structure & 3.02 & +0.171 \\
\midrule
Intercept & --- & +2.09 \\
\bottomrule
\end{tabular}
\end{table}

Table~\ref{tab:hd_categorical} presents results with categorical $N$ (R$^2 = 70.88\%$). The minimal improvement ($\Delta R^2 = +0.89\%$) indicates nearly linear size dependence. Critically, all features maintain their importance (ratios $\approx 0.99$), confirming they track true algorithmic state independent of problem scale. DFS depth, newly enabled nodes, and frontier size exhibit ratios of 0.99, demonstrating robustness to $N$ specification.

\begin{table}[h]
\centering
\caption{ANIRA-O feature importance with categorical $N$ fixed effects. High ratios ($\approx 0.99$) indicate features capture algorithmic state rather than proxying for graph size.}
\label{tab:hd_categorical}
\begin{tabular}{lrr}
\toprule
Feature & $\Delta R^2$ (\%) & Ratio \\
\midrule
DFS depth & 5.12 & 0.99 \\
Newly enabled & 3.72 & 0.99 \\
Frontier size & 3.60 & 0.99 \\
Hub structure & 2.85 & 0.94 \\
\bottomrule
\end{tabular}
\end{table}

\subsubsection{Model Comparison}

Table~\ref{tab:comparison} summarizes the categorical $N$ validation. ANIRA-E's large improvement ($+4.71\%$) reveals substantial non-linear size effects, while ANIRA-O's small gain ($+0.89\%$) indicates size scaling is approximately linear. This fundamental difference reflects distinct learned strategies: ANIRA-E exploits structural correlations with problem difficulty (dominated by graph size and hub structure), while ANIRA-O tracks algorithmic execution state (balanced importance across DFS depth, frontier size, and newly enabled nodes).

\begin{table}[h]
\centering
\caption{Comparison of linear vs categorical $N$ specifications. The magnitude of R$^2$ improvement when using categorical $N$ reveals the degree of non-linear size dependence in each model's compute allocation strategy.}
\label{tab:comparison}
\begin{tabular}{lrrr}
\toprule
Model & R$^2$ (Linear $N$) & R$^2$ (Categorical $N$) & Improvement \\
\midrule
ANIRA-E & 68.66\% & 73.37\% & $+4.71\%$ \\
ANIRA-O & 70.00\% & 70.88\% & $+0.89\%$ \\
\bottomrule
\end{tabular}
\end{table}

\subsubsection{Discussion}

The categorical $N$ analysis reveals qualitatively different learned behaviors despite similar task performance. ANIRA-E's compute allocation strategy relies primarily on structural heuristics, with graph size contributing 28\% of explained variance (linear model) and showing strong non-linear effects. Hub structure maintains moderate importance (3.51\%) when controlling for size, indicating it captures genuine difficulty signals, but algorithmic state features contribute minimally.

In contrast, ANIRA-O exhibits balanced feature importance with no dominant predictor. The algorithmic state features---DFS depth, frontier size, newly enabled nodes---maintain their explanatory power when controlling for graph size (ratios $\approx 0.99$), demonstrating they capture true execution state rather than structural proxies. The minimal categorical $N$ improvement ($+0.89\%$) indicates ANIRA-O's compute allocation scales approximately linearly with problem size, potentially offering superior generalization to novel graph sizes.

These findings demonstrate that architectural choices (predicting depth vs deciding when to halt) induce fundamentally different learned strategies. ANIRA-E learns to estimate problem difficulty from structure and allocate compute accordingly, while ANIRA-O learns to track algorithmic progress and decide when computation is sufficient. The latter's reliance on execution state features that directly correspond to the underlying DFS algorithm suggests better interpretability and more robust generalization properties.

\end{document}